\documentclass[10pt,twocolumn,letterpaper]{article}

\usepackage{iccv}
\usepackage{times}
\usepackage{epsfig}
\usepackage{graphicx}
\usepackage{amsmath}
\usepackage{amssymb}
\usepackage{multirow}
\usepackage{booktabs} 
\usepackage{subcaption}
\usepackage{balance}

\newcommand{\spin}{SPIN\xspace}

\newcommand\blfootnote[1]{%
  \begingroup
  \renewcommand\thefootnote{}\footnote{#1}%
  \addtocounter{footnote}{-1}%
  \endgroup
}

\usepackage[pagebackref=true,breaklinks=true,letterpaper=true,colorlinks,bookmarks=false]{hyperref}

\iccvfinalcopy 

\def\iccvPaperID{****} 

\ificcvfinal\fi
\begin{document}

\title{Learning to Reconstruct 3D Human Pose and Shape \\
via Model-fitting in the Loop}

\author{Nikos Kolotouros\textsuperscript{*}$^1$, Georgios Pavlakos\textsuperscript{*}$^1$, Michael J. Black$^2$, Kostas Daniilidis$^1$ \\[0ex]
$^1$ University of Pennsylvania \hspace{0.1em} $^2$ Max Planck Institute for Intelligent Systems
}

\maketitle
\ificcvfinal\thispagestyle{empty}\fi

\begin{abstract}
Model-based human pose estimation is currently approached through two different paradigms. Optimization-based methods fit a parametric body model to 2D observations in an iterative manner, leading to accurate image-model alignments, but are often slow and sensitive to the initialization. In contrast, regression-based methods, that use a deep network to directly estimate the model parameters from pixels, tend to provide reasonable, but not pixel accurate, results while requiring huge amounts of supervision. In this work, instead of investigating which approach is better, our key insight is that the two paradigms can form a strong collaboration. A reasonable, directly regressed estimate from the network can initialize the iterative optimization making the fitting faster and more accurate. Similarly, a pixel accurate fit from iterative optimization can act as strong supervision for the network. This is the core of our proposed approach SPIN (SMPL oPtimization IN the loop). The deep network initializes an iterative optimization routine that fits the body model to 2D joints within the training loop, and the fitted estimate is subsequently used to supervise the network. Our approach is self-improving by nature, since better network estimates can lead the optimization to better solutions, while more accurate optimization fits provide better supervision for the network. We demonstrate the effectiveness of our approach in different settings, where 3D ground truth is scarce, or not available, and we consistently outperform the state-of-the-art model-based pose estimation approaches by significant margins. The project website with videos, results, and code can be found at {\footnotesize \url{https://seas.upenn.edu/~nkolot/projects/spin}}.
\blfootnote{$^*$ equal contribution}   
\end{abstract}

\section{Introduction}
\begin{figure}
	\begin{subfigure}[]{.31\columnwidth}
		\includegraphics[width=\textwidth]{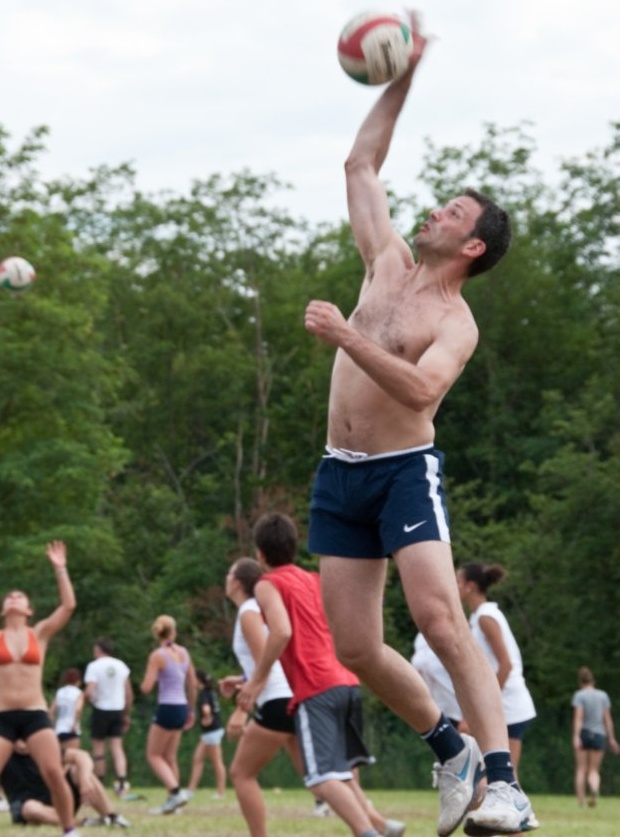}
	\end{subfigure}~
	\begin{subfigure}[]{.31\columnwidth}
		\includegraphics[width=\textwidth]{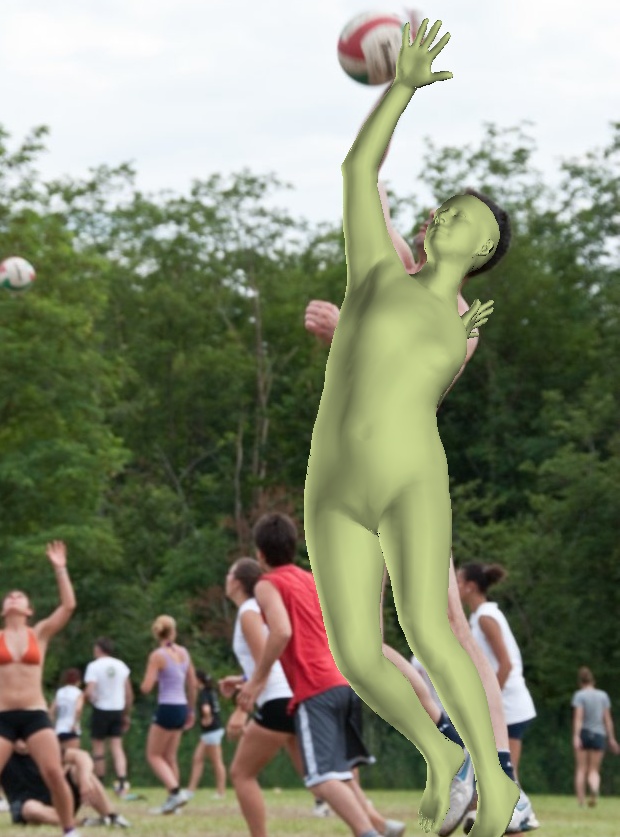}
	\end{subfigure}~
	\begin{subfigure}[]{.31\columnwidth}
		\includegraphics[width=\textwidth]{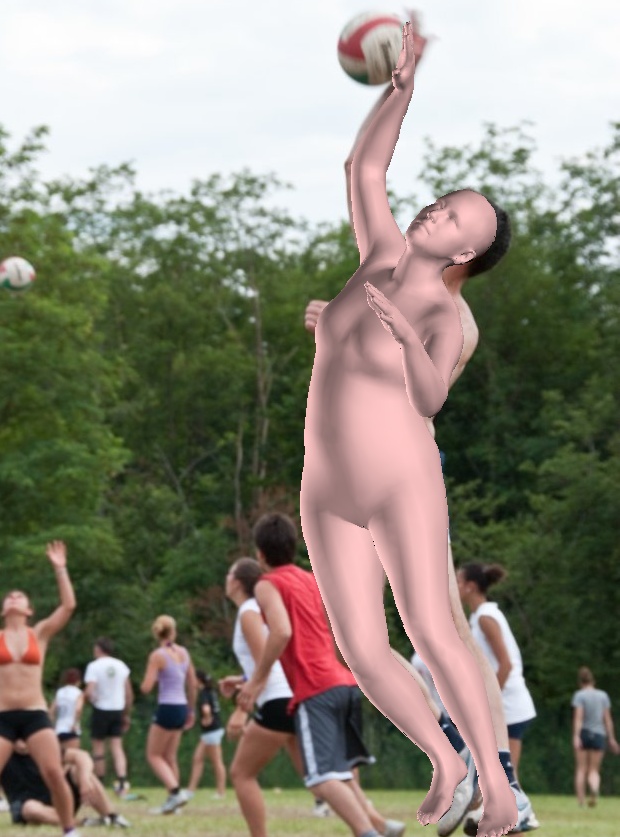}
	\end{subfigure}\\
	\begin{subfigure}[]{.31\columnwidth}
		\includegraphics[width=\textwidth]{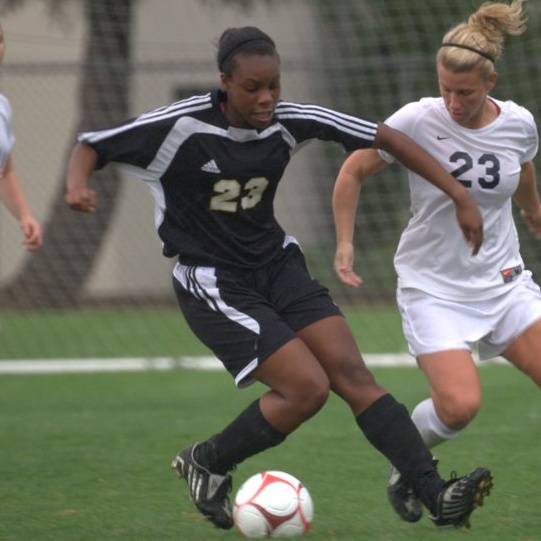}
			\caption*{Input image}		
	\end{subfigure}~
	\begin{subfigure}[]{.31\columnwidth}
		\includegraphics[width=\textwidth]{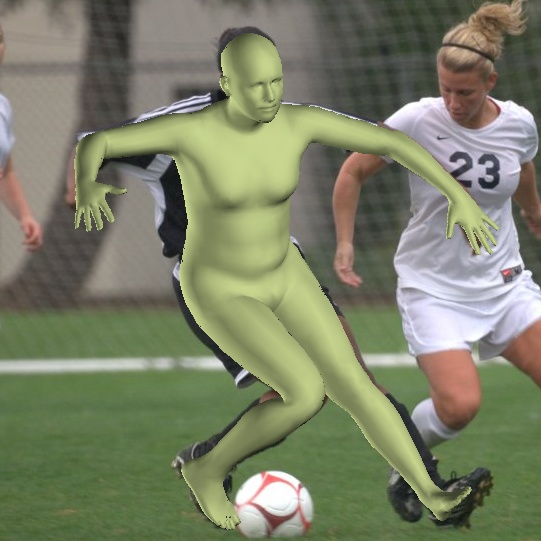}
			\caption*{Optimization result}		
	\end{subfigure}~
	\begin{subfigure}[]{.31\columnwidth}
		\includegraphics[width=\textwidth]{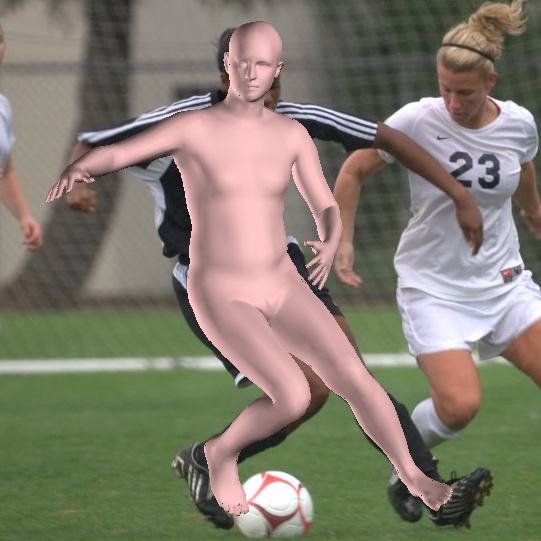}
			\caption*{Regression result}		
	\end{subfigure}
	\vspace{-2mm}
	\caption{Both optimization and regression approaches have successes and failures, so this motivates our approach to build a tight collaboration between the two.}
\label{fig:teaser}
\vspace{-2mm}
\end{figure}

\begin{figure*}[!t]
	\centering
	\includegraphics[scale=0.6]{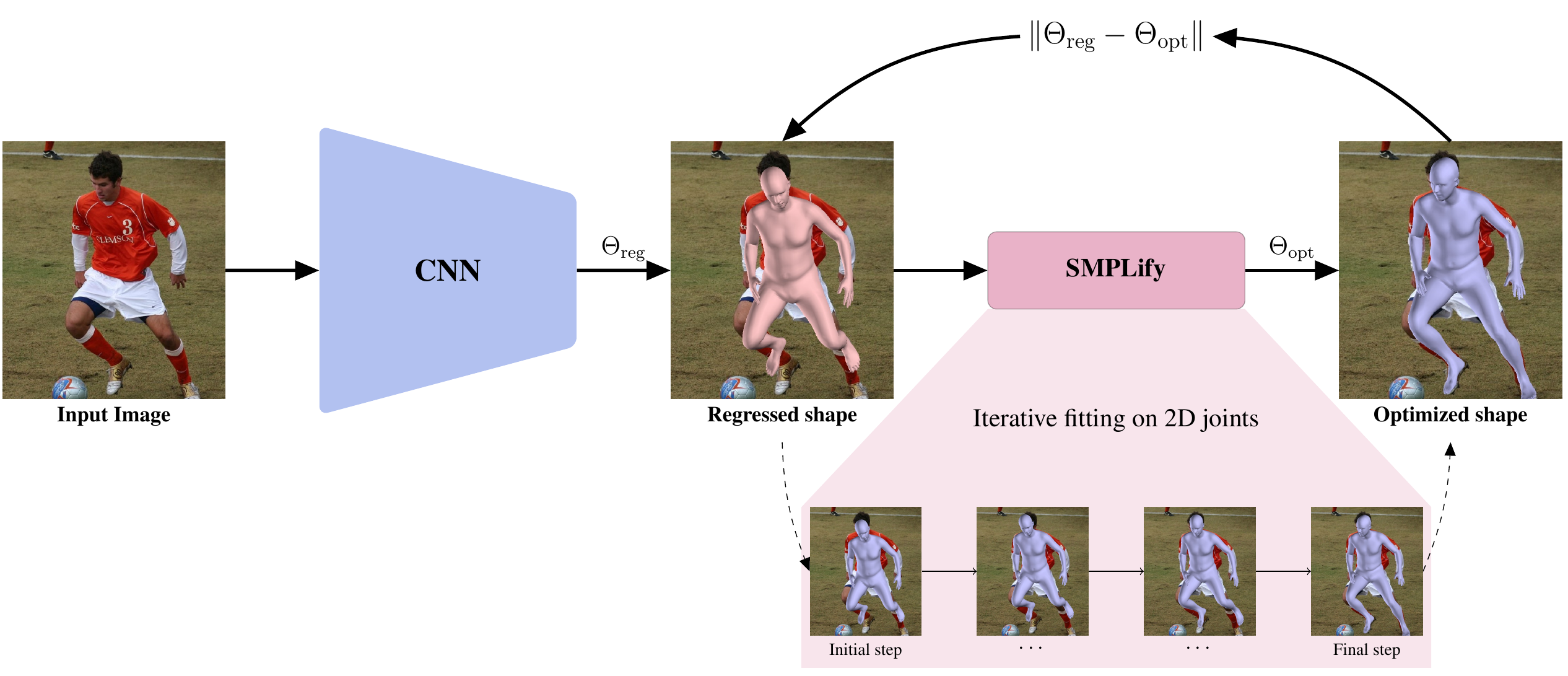}
	\vspace{-2mm}
	\caption{Overview of the proposed approach. \spin trains a deep network for 3D human pose and shape estimation through a tight collaboration between a regression-based and an iterative optimization-based approach. During training, the network predicts the parameters $\Theta_{reg}$ of the SMPL parametric model~\cite{loper2015smpl}. Instead of using the ground truth 2D keypoints to apply a weak reprojection loss, we instead propose to use our regressed estimate to initialize an iterative optimization routine that fits the model to 2D keypoints (SMPLify). This procedure is done {\em within the training loop}. The optimized model parameters $\Theta_{opt}$ are used to explicitly supervise the output of the network and supply it with privileged model-based supervision, that is beneficial compared to the weaker and typically ambiguous 2D reprojection losses. This collaboration leads to a self-improving loop, since better fits help the network train better, while better initial estimates from the network help the optimization routine converge to better fits.}
\label{fig:main}
\vspace{-2mm}
\end{figure*}

With the emergence of deep learning architectures, the dilemma between regression-based and optimization-based approaches for many computer vision problems has been more relevant than ever. Should we regress the relative camera pose, or use bundle adjustment? Is it more appropriate to regress the parameters of a face model, or fit the model to facial landmarks? These types of questions are ubiquitous within our community. Among others, 3D model-based human pose estimation has initiated similar discussions, since both optimization-based~\cite{bogo2016keep,lassner2017unite} and regression-based approaches~\cite{kanazawa2018end,omran2018neural,pavlakos2018learning} have had significant success recently. However, one can argue that both paradigms have weak and strong points (Figure~\ref{fig:teaser}). Based on this, in this work we advocate that instead of focusing on which paradigm is better, if we aim to push the field forward, we need to consider ways for collaboration between the two.

Although 3D model-based human pose is a very challenging and highly ambiguous problem, there have been fundamental works that attempt to address it. Optimization-based methods~\cite{bogo2016keep,guan2009estimating,lassner2017unite}, are pretty well explored and understood. Given a parametric model of the human body, e.g., SMPL~\cite{loper2015smpl}, an iterative fitting approach attempts to estimate the body pose and shape that best explains 2D observations, most typically 2D joint locations. Since we explicitly optimize for the agreement of the model with image features, we typically get a good fit, but the optimization tends to be very slow and is quite sensitive to the choice of the initialization. On the other hand, recent deep learning advances have shifted the spotlight towards purely regression-based methods, using deep networks to regress the parameters of the model directly from images~\cite{kanazawa2018end,omran2018neural,pavlakos2018learning}. In theory, this is a very promising direction, since the deep regressor can take all pixel values into consideration, instead of relying only on a sparse set of 2D locations. Unfortunately, this type of one-shot prediction might lead to mediocre image-model alignment, while at the same time a large amount of data is necessary to properly train the network. So naturally, there is a large list of arguments in favor and against each method.

In this work, we advocate that instead of arguing over one paradigm or the other we should embrace the strengths and the weaknesses of each method and use them in a tight collaboration during training. In our approach, a deep network is used to regress the parameters of the SMPL parametric model~\cite{loper2015smpl}. These regressed values initialize the iterative fitting routine that aligns the model to the image given the 2D keypoints. Subsequently, the parameters of the fitted model are used as supervision for the network, closing the loop between the regression and the optimization method. This is the core of our approach, \spin, that fits the model within the training loop, and uses it as a privileged form of supervision for the neural network (Figure~\ref{fig:main}). A critical characteristic of our proposed approach is that it is self-improving by nature. In the early training stages, the network will produce results close to the mean pose meaning that the iterative fitting will be prone to make errors. As more examples are provided to the network as supervision by the iterative fitting module, it will learn to produce more meaningful shapes that will also lead the optimization to more accurate model fits. Moreover, since the iterative fitting requires only 2D keypoints to fit the model, our network can be trained even when no image with corresponding 3D ground truth is available, since the 3D supervision will be provided by the optimization module. Finally, and most crucially in terms of performance, our network is trained with explicit 3D supervision, in the form of model parameters and full shape instead of weaker 2D reprojection errors as in previous works~\cite{kanazawa2018end,pavlakos2018learning}. This privileged form of supervision turns out to be very important to improve the regression performance. Our approach is benchmarked in different settings and in a variety of indoor and in-the-wild datasets and it outperforms state-of-the-art model-based approaches by a significant margin.

We summarize the contributions of our approach below:
\begin{itemize}
\item We present \spin, a self-improving approach for training a neural network for 3D human pose and shape estimation, through the tight collaboration of a regression- and an optimization-based method.
\item Since the supervision is supplied by the iterative fitting module, training is feasible even when no image with 3D ground truth is available for training.
\item The fitted model supplies our network with explicit model-based supervision which is crucial to improve performance compared to weaker 2D supervision (e.g., reprojection losses).
\item We achieve state-of-the-art results in model-based 3D pose and shape estimation across many benchmarks.
\end{itemize}

\section{Related work}
Recent works have made significant advances in the frontier of skeleton-based 3D human pose estimation from single images, with many approaches achieving impressive results~\cite{martinez2017simple,mehta2017vnect,rogez2019lcr,sun2018integral,tekin2017learning,zhou2016sparseness}. Although this line of work has boosted the interest for 3D human pose estimation, here we will focus our review on model-based pose estimation. Approaches in this category consider a parametric model of the human body, like SMPL~\cite{loper2015smpl} or SCAPE~\cite{anguelov2005scape}, and the goal is to estimate the full body 3D pose and shape.

\textbf{Optimization-based methods}:
Optimization-based approaches used to be the leading paradigm for model-based human pose estimation. Early work in the area~\cite{guan2009estimating, sigal2008combined} attempted to estimate the parameters of the SCAPE model using silhouettes or keypoints and often there was some manual user intervention needed. Recently, the first fully automatic approach, SMPLify, was introduced by Bogo~\etal~\cite{bogo2016keep}. Using an off-the-shelf keypoint detector~\cite{pishchulin2016deepcut}, SMPLify fits SMPL to 2D keypoint detections, using strong priors to guide the optimization. Beyond SMPLify, different updates to the standard pipeline have investigated incorporating in the fitting procedure, silhouette cues~\cite{lassner2017unite}, multiple views~\cite{huang2017towards}, or even handle multiple people~\cite{zanfir2018monocular}. More recently, works have demonstrated fits for more expressive models in the multi-view~\cite{joo2018total}, as well as the single-view setting~\cite{pavlakos2019expressive,xiang2019monocular}. In this work, we exploit the particular effectiveness of optimization-based approaches to produce pixel-accurate fittings, but instead of using them to produce good predictions at test time, our goal is to leverage them to supply direct supervision for a neural network.

\textbf{Regression-based methods}:
On the other end of the spectrum, recent works rely exclusively on regression to address the problem of 3D human pose and shape estimation. In most cases, given a single RGB image, a deep network is used to regress the model parameters. Considering the lack of images with full 3D shape ground truth, the majority of these works have focused on alternative supervision signals to train the deep networks. Most of them rely heavily on 2D annotations including 2D keypoints, silhouettes, or parts segmentation. This information can be used as input~\cite{tung2017self}, intermediate representation~\cite{omran2018neural,pavlakos2018learning}, or as supervision, by enforcing different reprojection losses~\cite{kanazawa2018end,omran2018neural,pavlakos2018learning,tan2017indirect,tung2017self}. Although these constraints are very useful, they are providing weak supervision for the network. Instead, we argue that strong model-based supervision, i.e., direct supervision on the model parameters and/or output mesh is crucial to improve performance. Although this type of ground truth is rarely available, we use a fitting routine in the training loop to provide the strong supervision signal to train the network. 

\textbf{Iterative fitting meets direct regression}:
Ideas of using regression approaches to improve fitting and vice versa have also been considered before in the literature. Early optimization methods required a good initial estimate which could be obtained by a discriminative approach~\cite{sigal2008combined}. Lassner~\etal~\cite{lassner2017unite} used SMPLify to get good model fits, which could be later used for regression tasks (e.g., part segmentation or landmark detection). Rogez~\etal~\cite{rogez2019lcr} also employed 3D pose pseudo annotations for training. Pavlakos~\etal~\cite{pavlakos2018learning} used an initial prediction from their network to initialize and anchor the SMPLify optimization routine. Varol~\etal~\cite{varol2018bodynet} proposed an extension of SMPLify to fit SMPL on the regressed volumetric representation of their network. Although previous works have also considered the benefits of these two approaches, in our work we propose a much tighter collaboration by incorporating the fitting method within the training loop, in a self-improving manner, to harness better supervision for the network.

To put our approach in a larger context, the idea of combining direct regression networks with different optimization routines has also emerged in different settings. Training a network jointly with a graphical model has been proposed by Tompson~\etal~\cite{tompson2014joint} in the context of 2D human pose estimation. Similarly, for segmentation, it is popular to use a CRF on top of the segmentation network~\cite{chen2018deeplab}, while, unrolling the CRF optimization to train the network jointly with the optimization has also been investigated~\cite{schwing2015fully,zheng2015conditional}. These ideas have also translated to 3D, where Paschalidou~\etal~\cite{paschalidou2018raynet} unrolls the MRF optimization to train it jointly with a network for depth regression. Although we draw inspiration from these works, our motivation is different since instead of unrolling the optimization, or doing a simple post-processing, we leverage the iterative fitting to provide strong supervision to the network.

\section{Technical approach}
In the following, we describe the parametric human body model, SMPL~\cite{loper2015smpl}, and we define the basic notation. Then we provide more details about the regression network and the iterative optimization routine, based on SMPLify~\cite{bogo2016keep}. Finally, we describe our approach, \spin, and give the necessary implementation details.

\subsection{SMPL model}
The SMPL body model~\cite{loper2015smpl}, provides a function $\mathcal{M}(\theta, \beta)$ that takes as input the pose parameters $\theta$ and the shape parameters $\beta$, and returns the body mesh $M \in \mathbb{R}^{N\times 3}$, with $N=6890$ vertices. Conveniently, the body joints $X$ of the model can be defined as a linear combination of the mesh vertices. A linear regressor $W$ can be pre-trained for this task, so for $k$ joints of interest, we define the major body joints $X \in \mathbb{R}^{k \times 3} = WM$.

\subsection{Regression network}
For the regression model, we use a deep neural network. Our architecture has the same design with Kanazawa~\etal~\cite{kanazawa2018end} with the only difference that we use the representation proposed by Zhou~\etal~\cite{zhou2018continuity} for the 3D rotations, since we empirically observed faster convergence during training. Let us now denote with $f$ the function approximated by the neural network. A forward pass of a new image provides the regressed prediction for the model parameters $\Theta_{reg} = \{\theta_{reg}, \beta_{reg}\}$ and the camera parameters $\Pi_{reg}$. These parameters allow us to estimate the 2D projection of the joints $J_{reg} = \Pi_{reg}(X_{reg})$. Our prediction allows us to generate the mesh corresponding to the regressed parameters, $M_{reg} = \mathcal{M}(\theta_{reg}, \beta_{reg})$, as well as the joints and their reprojection $J_{reg}$. In this setting, a common supervision is provided using a reprojection loss on the joints:
\begin{equation}
L_{\text{2D}} =  ||{J}_{reg} - {J}_{gt}||,
\end{equation}
where ${J}_{gt}$ are the ground truth 2D joints. However, in this work, we argue that this supervisory signal is very weak and puts an extra burden on the network, forcing it to search in the parameter space for a valid pose that agrees with the ground truth 2D locations.

\subsection{Optimization routine}
The iterative fitting routine follows the SMPLify work by Bogo~\etal~\cite{bogo2016keep}. We give a short introduction here, but we also refer the reader to~\cite{bogo2016keep} for more details. SMPLify tries to fit the SMPL model to a set of 2D keypoints using an optimization-based approach. The objective function it minimizes consists of a reprojection loss term and a number of pose and shape priors. More specifically, the total objective is:
\begin{equation}
E_J(\beta, \theta; K, J_{est}) + \lambda_\theta E_{\theta}(\theta) + \lambda_a E_{a}(\theta)  + \lambda_\beta E_\beta(\beta)
\label{eq:smplify_loss}
\end{equation}
where $\beta$ and $\theta$ are the parameters of the SMPL model, $J_{\mathrm{est}}$ the detected 2D joints and $K$ the camera parameters. The first term $E_J(\beta, \theta; K, J_{est})$ is a penalty on the weighted 2D distance between $J_{\mathrm{est}}$ and the projected SMPL joints. $E_{\theta}(\theta)$ is a mixture of Gaussians pose prior trained with shapes fitted on marker data, $E_{a}(\theta)$ is a pose prior penalizing unnatural rotations of elbows and knees, while $E_\beta(\beta)$ is a quadratic penalty on the shape coefficients. We did not include the interpenetration error term of~\cite{bogo2016keep}, since it makes fitting slower, while having little performance benefit.

The first step of SMPLify involves an optimization over the camera translation and body orientation, while keeping the model pose and shape fixed. After estimating the camera translation, SMPLify attempts to minimize \eqref{eq:smplify_loss}, using a 4-stage fitting procedure. The 4-stage optimization is crucial to avoid getting trapped in local minima because the optimization is initialized from the mean pose. In contrast, since our approach uses the network prediction to initialize the optimization, we observed that a single optimization stage, with a small number of iterations, is typically enough to converge to a good fit. Also instead of estimating the initial translation using triangle similarity as in~\cite{bogo2016keep} we can also use the predicted camera translation from the network. This can be helpful in cases where the assumptions made in~\cite{bogo2016keep} (e.g., person is always standing) are not valid.  

Another modification aiming at faster runtime is that we run SMPLify in batch mode. Instead of optimizing for each image sequentially, the optimization runs in parallel. Although SMPLify can have high latency that makes it unsuitable for single-image inference, we can achieve high throughput on a modern GPU by optimizing for several examples concurrently. Moreover, while SMPLify uses joints $J_{est}$ along with their detection confidences provided by DeepCut~\cite{pishchulin2016deepcut}, for our ground truth, we can only assume that all joints have the same confidence. This can affect negatively the fitting procedure, since typically there are small annotation mistakes, e.g., annotating joints under occlusion, or generally geometrically inconsistent annotations. To alleviate this problem, we combine the provided ground truth 2D joints for each person with the corresponding OpenPose detections~\cite{cao2018openpose,cao2017realtime,simon2017hand,wei2016convolutional}. This enables us to leverage the confidence in each detection and avoid mistakes because of high-confidence erroneous annotations.

\begin{figure}[!t]
	\centering
	\includegraphics[scale=0.32]{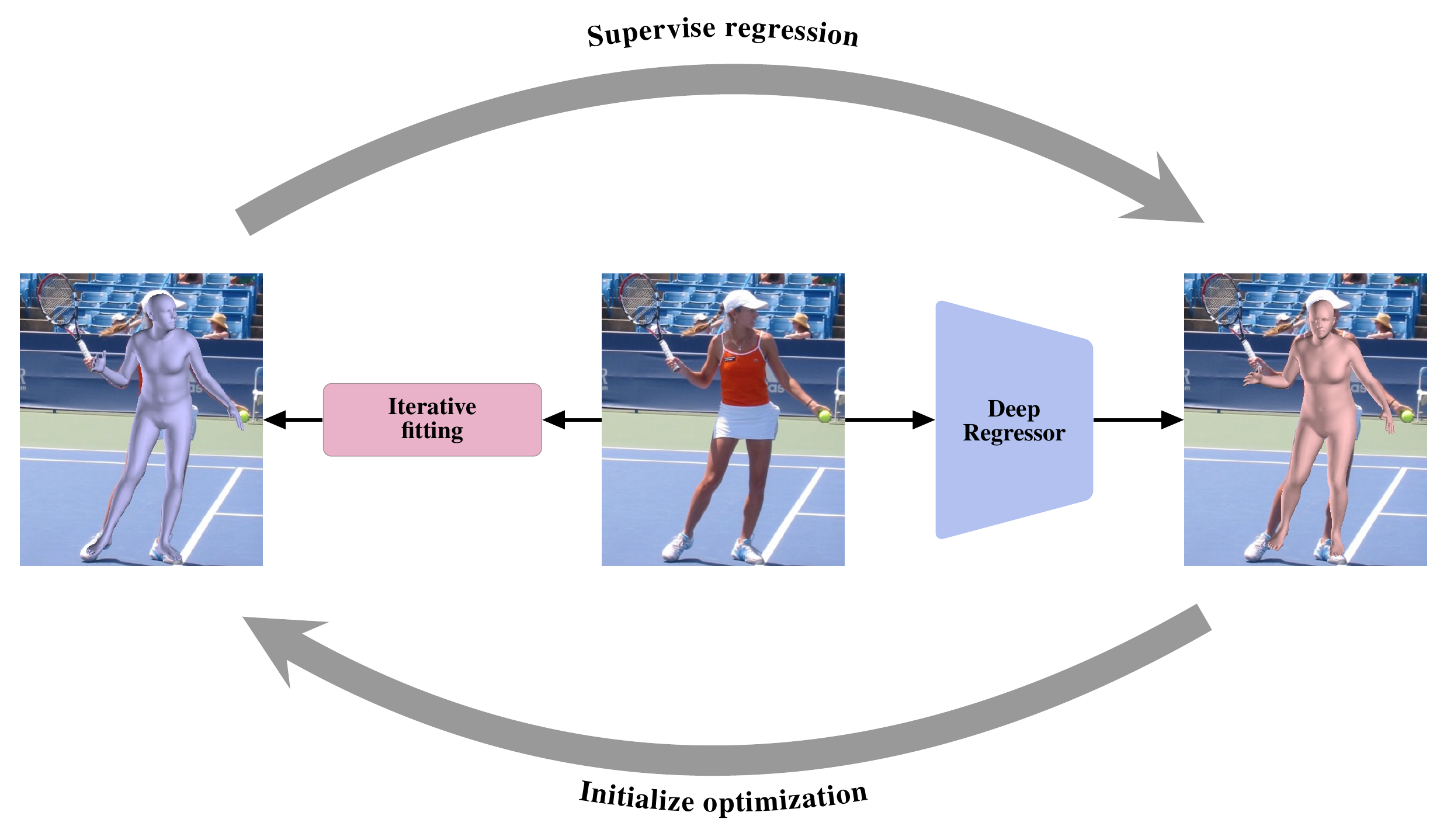}
	\vspace{-2mm}
	\caption{
\spin builds a tight collaboration between an optimization-based and a regression-based approach. A reasonable regressed estimate from the network initializes properly the optimization, thus leading to a better optimum. Similarly, a value optimized by iterative fitting can act as supervision to better train the network. The two procedures continue this collaboration forming a self-improving loop.}
\label{fig:spin}
\vspace{-2mm}
\end{figure}

\subsection{\spin}
Our approach, \spin, builds on the insight that the previous two paradigms can form a tight collaboration to train a deep regressor for human pose and shape estimation (Figure~\ref{fig:spin}). During a typical training loop, an image is forwarded through the network providing the regressed parameters $\Theta_{reg}$. Instead of applying the typical 2D reprojection losses right away, the regressed parameters are instead used to initialize the optimization routine. This optimization is usually very slow if we start from the mean pose as an initial value. However, given a reasonable initial estimate, it can be greatly accelerated. This enables us to employ the fitting routine within the training loop. Let us now denote with $\Theta_{opt} = \{\theta_{opt}, \beta_{opt}\}$ the set of model parameters produced by the iterative fitting. These values are explicitly optimized such that the produced shape $M_{opt} = \mathcal{M}(\theta_{opt}, \beta_{opt})$ and reprojected joints $J_{opt}$, align with the 2D keypoints. Given these optimized values, we can directly supervise the network function $f$ on the parameter level:
\begin{equation}
L_{\text{3D}} =  ||\Theta_{reg} - \Theta_{opt}||,
\end{equation}
and/or the mesh level:
\begin{equation}
L_{\text{M}} =  ||M_{reg} - M_{opt}||.
\end{equation}
In practice, this has a very different effect than applying a reprojection loss for the 2D joints. Instead of forcing the network to identify a set of parameters that satisfy the joints reprojection, we supply it directly with a parametric solution that corresponds to a feasible 3D shape. Intuitively, we bypass the search of the network on the parameter space, and we directly provide a privileged set of parameters $\Theta_{opt}$ which tend to be very close to the actual optimal solution.

Another crucial characteristic of \spin is that it is self-improving by nature. A good initial network estimate $\Theta_{reg}$ will lead the optimization to a better fit $\Theta_{opt}$, while a good fit from the iterative routine will provide even better supervision to the network. This makes running the routine in the loop particularly important, since it enables the close collaboration between the two components.

Moreover, since the optimization routine uses only 2D joints for the fitting, and the network relies primarily on this routine for the necessary model-based supervision, our approach is applicable even in cases where no image with corresponding 3D ground truth is available for training. This resembles the unpaired setting of~\cite{kanazawa2018end}, where only 2D keypoint annotations are available, and an adversarial prior is trained to penalize invalid poses/shapes. The benefit of our approach in this setting is that instead of providing a yes/no answer to the network as the discriminator does, we explicitly supervise it with a valid pose, which leads to better performance empirically, as we demonstrate in our evaluation.

\subsection{Implementation details}
Here we discuss in more detail some further implementation details that were important for the training procedure. Although SMPLify is quite accurate, for some cases we can still get bad failures. These bad fits can make training unstable and potentially decrease performance. This motivated us to use a criterion to reject supervision from these shapes. Empirically, a simple thresholding based on the joint reprojection error worked very well in our case. For the images with rejected fits, we only supervise the regression network with a reprojection loss on the joints. Additionally, to avoid training with improbable values for the shape parameters (i.e., beyond $\pm 3\sigma$), when SMPLify returns shape values outside this range, we only supervise the $\beta$ parameters with a simple $L_2$ loss, i.e., pushing it close to the mean shape.

To improve and accelerate training, we also incorporated a dictionary, such that for each image in our training set we can keep track of the best fit we have seen for it over all epochs. In practice, every time we compute a new optimized shape in the loop, we compare with the best fit we have seen until that point in time and if the new fit is better, we update the dictionary accordingly. To compare the quality of the fits, we again use the reprojection error on the joints. Our dictionary is initially populated with SMPLify fits, a process done offline before the training starts. To initialize SMPLify for this process, we can start from the mean pose, or use a more accurate pose, regressed from the 2D keypoints (e.g., using a network similar to Martinez~\etal~\cite{martinez2017simple}). For our empirical evaluation we focus on the second strategy, but we also present similar results with the first approach in the Sup.Mat. We run the SMPLify optimization for a total of 50 iterations for each batch. 

\section{Empirical evaluation}
\subsection{Datasets}
Here we give a quick description of the datasets we use for training and evaluation. We report results on Human3.6M~\cite{ionescu2014human3}, MPI-INF-3DHP~\cite{mehta2017monocular}, LSP~\cite{johnson2010clustered}, and 3DPW~\cite{von2018recovering}. We train using the first three datasets (no training data from 3DPW), while similarly to~\cite{kanazawa2018end}, we also incorporate training data with 2D annotations from other datasets, i.e., LSP-Extended~\cite{johnson2011learning}, MPII~\cite{andriluka20142d}, and COCO~\cite{lin2014microsoft}. For the different settings we investigate, e.g., training with/without in the loop update, or training with/without 3D ground truth), we train a single model per setting and we use it to report results on all datasets, without fine-tuning on each particular dataset. Moreover, we clarify, that we always evaluate the network's output. No additional fitting-based post-processing is applied, as is done for example in~\cite{guler2019holopose}. Also, since different datasets often use different error metrics to report results, we use the metrics that are more often met in the literature for each dataset. We give a detailed definition of the various metrics in Sup.Mat.

\noindent
\textbf{Human3.6M}:
It is an indoor benchmark for 3D human pose estimation. It includes multiple subjects performing actions like Eating, Sitting and Walking. Following typical protocols, e.g.,~\cite{kanazawa2018end}, we use subjects S1, S5, S6, S7, S8 for training and we evaluate on subjects S9 and S11.

\noindent
\textbf{MPI-INF-3DHP}:
It is a dataset captured with a multi-view setup mostly in indoor environments. No markers are used for the capture, so 3D pose data tend to be less accurate compared to other datasets. We use the provided training set (subjects S1 to S8) for training and we report results on the test set of the dataset.

\noindent
\textbf{LSP}:
It is a standard dataset for 2D human pose estimation. Here we employ the test set for evaluation, using the silhouette/parts annotations from Lassner~\etal~\cite{lassner2017unite}.

\noindent
\textbf{3DPW}:
It is a very recent dataset, captured mostly in outdoor conditions, using IMU sensors to compute pose and shape ground truth. We use this dataset only for evaluation on its defined test set.

\begin{figure}
    \begin{subfigure}[]{.31\columnwidth}
		\includegraphics[width=\textwidth]{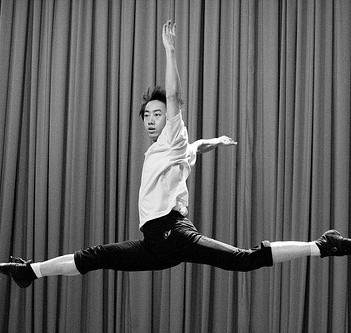}
	\end{subfigure}~
	\begin{subfigure}[]{.31\columnwidth}
		\includegraphics[width=\textwidth]{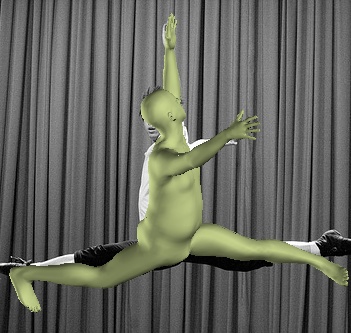}
	\end{subfigure}~
	\begin{subfigure}[]{.31\columnwidth}
		\includegraphics[width=\textwidth]{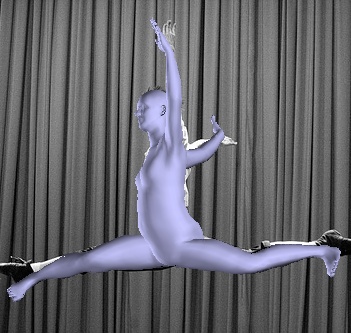}
	\end{subfigure}\\
	\begin{subfigure}[]{.31\columnwidth}
		\includegraphics[width=\textwidth]{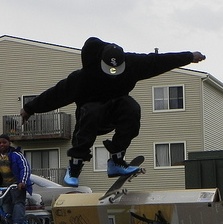}
	\end{subfigure}~
	\begin{subfigure}[]{.31\columnwidth}
		\includegraphics[width=\textwidth]{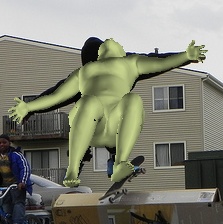}
	\end{subfigure}~
	\begin{subfigure}[]{.31\columnwidth}
		\includegraphics[width=\textwidth]{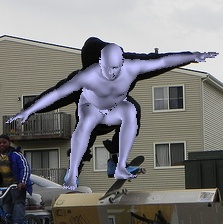}
	\end{subfigure}\\
	\begin{subfigure}[]{.31\columnwidth}
		\includegraphics[width=\textwidth]{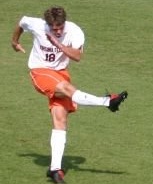}
	\end{subfigure}~
	\begin{subfigure}[]{.31\columnwidth}
		\includegraphics[width=\textwidth]{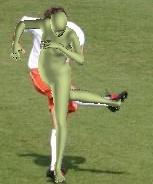}
	\end{subfigure}~
	\begin{subfigure}[]{.31\columnwidth}
		\includegraphics[width=\textwidth]{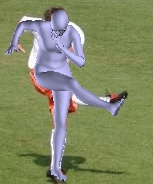}
	\end{subfigure}\\
	\begin{subfigure}[]{.31\columnwidth}
		\includegraphics[width=\textwidth]{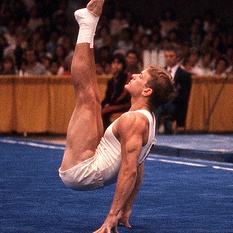}
			\caption*{Input image}		
	\end{subfigure}~
	\begin{subfigure}[]{.31\columnwidth}
		\includegraphics[width=\textwidth]{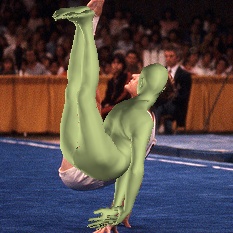}
			\caption*{Initial fit}		
	\end{subfigure}~
	\begin{subfigure}[]{.31\columnwidth}
		\includegraphics[width=\textwidth]{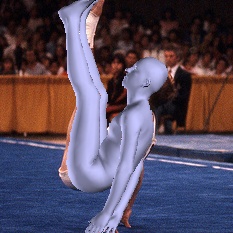}
			\caption*{Final fit}
	\end{subfigure}
	\vspace{-2mm}
	\caption{Examples of SMPLify fits in our dictionary at the beginning of training and at the end of training.
Although SMPLify can fail when starting from an inaccurate pose (second column), given a good prediction from our network as initialization, the optimization can converge to an accurate solution (third column).}
\label{fig:comparison}
\vspace{-2mm}
\end{figure}

\subsection{Quantitative evaluation}
\textbf{Ablative studies}:
First we evaluate the components of our approach. We use in-the-wild datasets for this, since they are much more challenging, compared to the indoor benchmarks, where the models tend to overfit~\cite{ionescu2014human3,mehta2017monocular}.

\begin{table}
\centering
\small
\hspace{-3mm}
\tabcolsep=0.85mm
\begin{tabular}{@{}lc@{}}
\toprule
& Rec. Error \\
\midrule
HMR~\cite{kanazawa2018end} & 81.3 \\
Kanazawa~\etal~\cite{kanazawa2018dynamics} & 72.6 \\
Arnab~\etal~\cite{arnab2019exploiting} & 72.2 \\
Kolotouros~\etal~\cite{kolotouros2019convolutional} & 70.2 \\
Ours - static fits & 66.3 \\
Ours - in the loop & \bf{59.2} \\
\bottomrule
\end{tabular}
\vspace{-2mm}
\caption{Evaluation on the 3DPW dataset. The numbers are mean reconstruction errors in mm. The model-based supervision alone (Ours - static fits) outperforms similar architectures trained on the same (\cite{kanazawa2018end,kolotouros2019convolutional}) or more data (\cite{arnab2019exploiting,kanazawa2018dynamics}). Incorporating the fitting in the loop (Ours - in the loop) further improves performance.}
\label{tab:3dpw}
\vspace{-2mm}
\end{table}

On the new 3DPW dataset, we evaluate pose estimation. In Table~\ref{tab:3dpw}, we provide the results for two versions of our approach, one where the network is supervised only with the initial dictionary fits, without running the optimization in the loop (Ours - static fits), and a second where we run the optimization in the loop, and the network can benefit from the improved fits that the iterative fitting tends to produce (Ours - in the loop). To put our results into perspective, we also compare with four recent baselines (\cite{arnab2019exploiting,kanazawa2018end,kanazawa2018dynamics,kolotouros2019convolutional}). As we can see, the use of model supervision is enough to improve performance over the other baselines. Unsurprisingly, running the iterative fitting in the loop, we can further improve the performance of the network, since it gradually gets access to better and better fits.

\begin{table}
\centering
\small
\hspace{-3mm}
\tabcolsep=2.95mm
\begin{tabular}{@{}lcccc@{}}
\toprule
& \multicolumn{2}{c}{FB Seg.} & \multicolumn{2}{c}{Part Seg.} \\
\cmidrule{2-5}
& acc. & f1 & acc. & f1 \\
\midrule
SMPLify \emph{oracle} & \bf{92.17} &  \bf{0.88} & 88.82 & 0.67 \\
SMPLify & 91.89 &  \bf{0.88} & 87.71 & 0.64 \\
SMPLify on~\cite{pavlakos2018learning} & 92.17 &  \bf{0.88} & 88.24 & 0.64 \\
HMR~\cite{kanazawa2018end} & 91.67 & 0.87 & 87.12 & 0.60 \\
Ours - static fits & 91.07 & 0.86 & 88.48 & 0.65 \\
Ours - in the loop & 91.83 & 0.87 & \bf{89.41} & \bf{0.68} \\
\bottomrule
\end{tabular}
\vspace{-2mm}
\caption{Evaluation on foreground-background and six-part segmentation on the LSP test set. The numbers are accuracies and f1 scores. Using the model-based supervision without updating the fits achieves very competitive results, while the incorporation of the fitting in the loop propels our approach beyond the state-of-the-art. The numbers for the first two rows are taken from~\cite{lassner2017unite}.}
\label{tab:lsp}
\vspace{-2mm}
\end{table}

\begin{figure*}[!t]
	\centering
	\begin{subfigure}[]{.15\textwidth}
		\includegraphics[width=\textwidth]{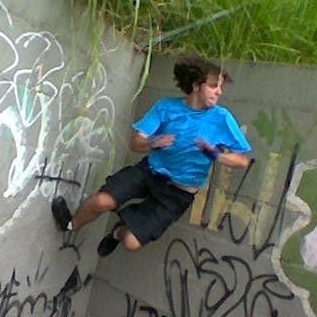}
	\end{subfigure}~
	\begin{subfigure}[]{.15\textwidth}
		\includegraphics[width=\textwidth]{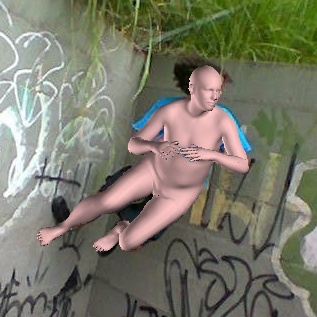}
	\end{subfigure}~
	\begin{subfigure}[]{.15\textwidth}
		\includegraphics[width=\textwidth]{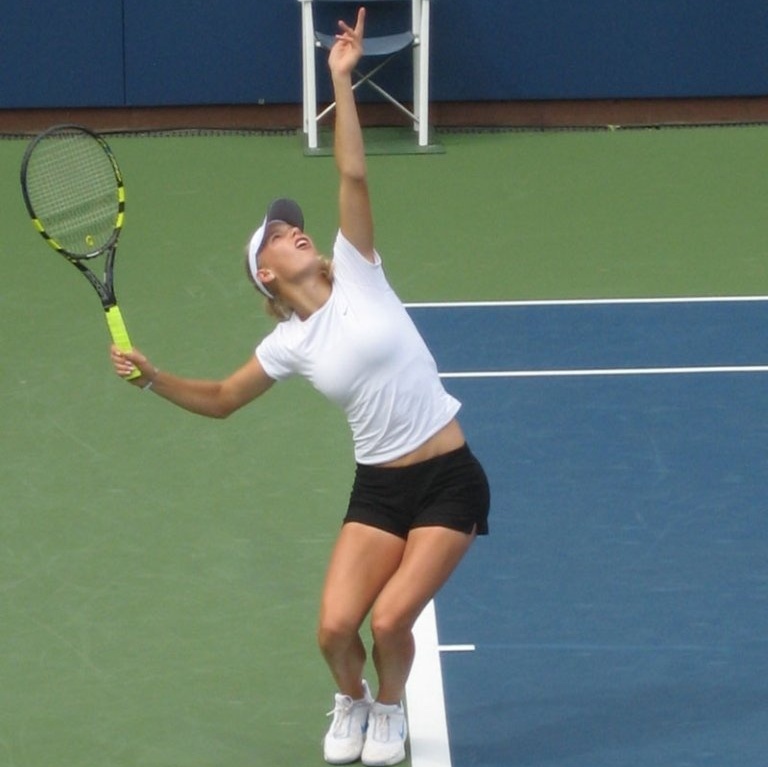}
	\end{subfigure}~
	\begin{subfigure}[]{.15\textwidth}
		\includegraphics[width=\textwidth]{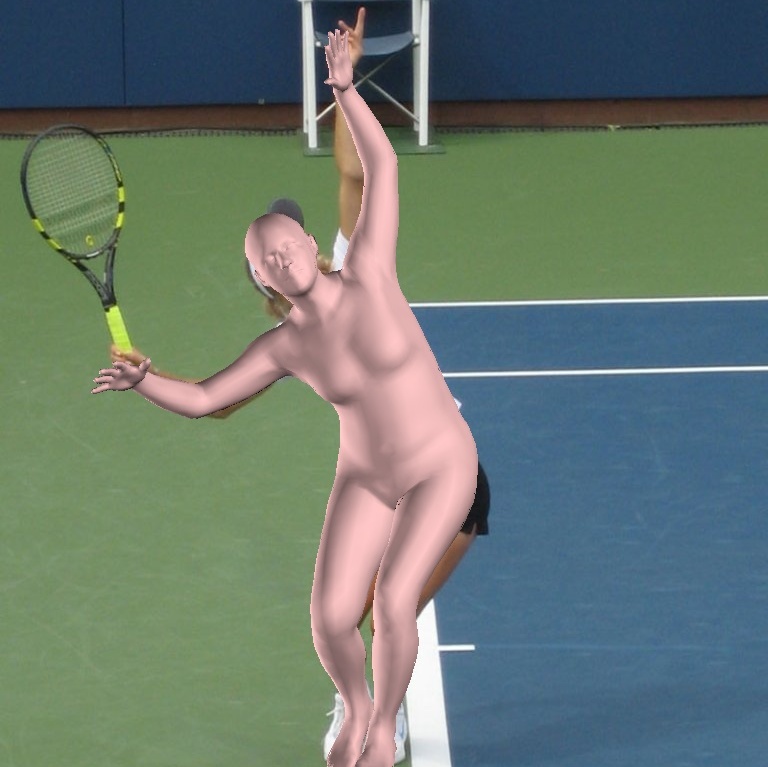}
	\end{subfigure}~
	\begin{subfigure}[]{.15\textwidth}
		\includegraphics[width=\textwidth]{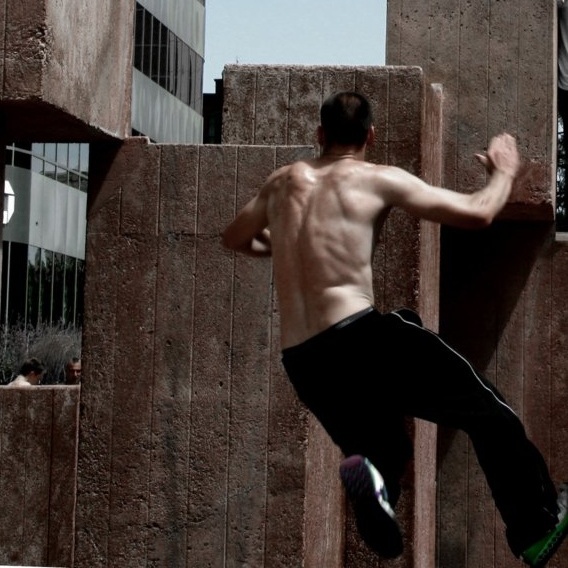}
	\end{subfigure}~
	\begin{subfigure}[]{.15\textwidth}
		\includegraphics[width=\textwidth]{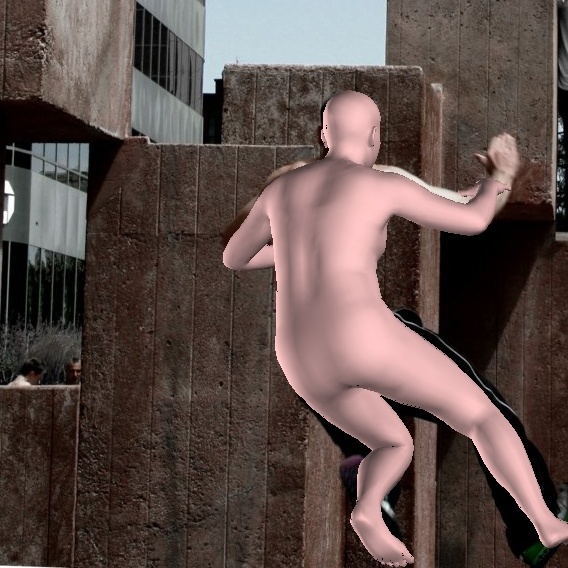}
	\end{subfigure}\\
	\begin{subfigure}[]{.15\textwidth}
		\includegraphics[width=\textwidth]{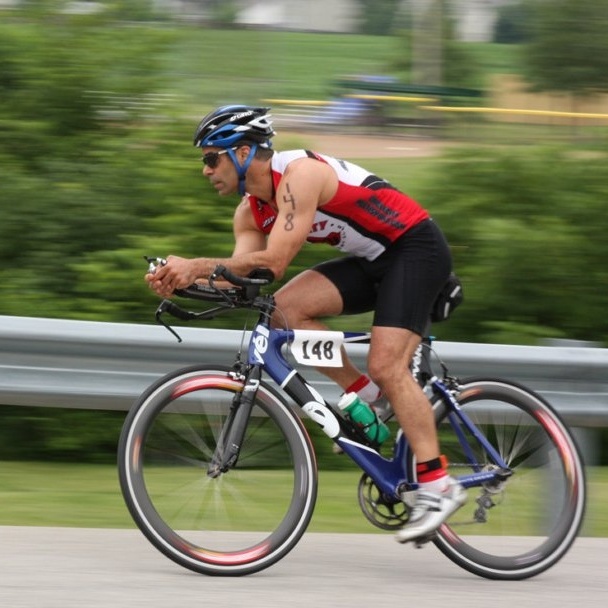}
	\end{subfigure}~
	\begin{subfigure}[]{.15\textwidth}
		\includegraphics[width=\textwidth]{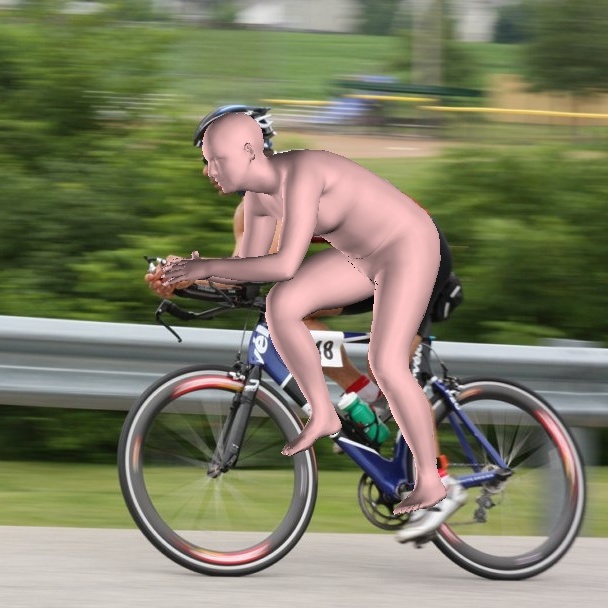}
	\end{subfigure}~
	\begin{subfigure}[]{.15\textwidth}
		\includegraphics[width=\textwidth]{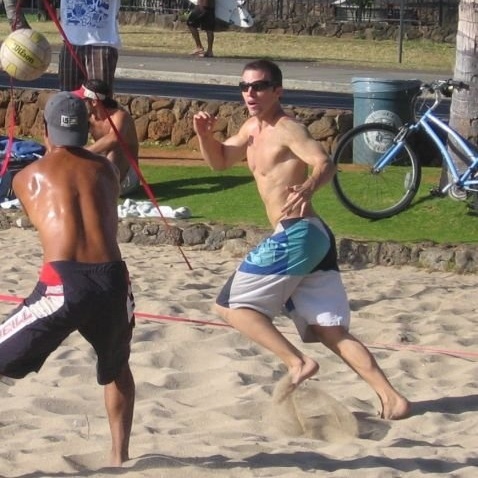}
	\end{subfigure}~
	\begin{subfigure}[]{.15\textwidth}
		\includegraphics[width=\textwidth]{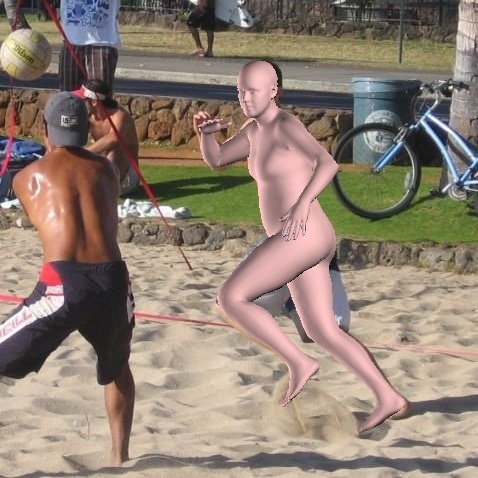}
	\end{subfigure}~
	\begin{subfigure}[]{.15\textwidth}
		\includegraphics[width=\textwidth]{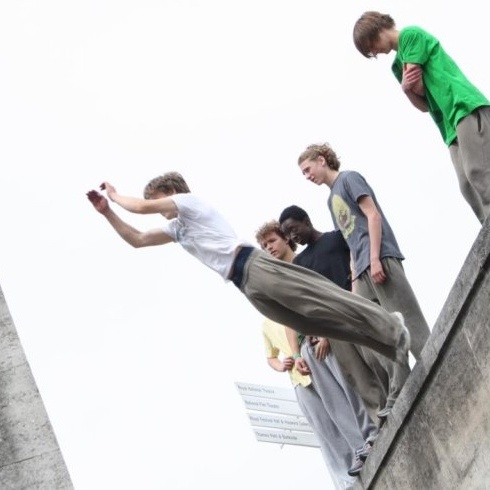}
	\end{subfigure}~
	\begin{subfigure}[]{.15\textwidth}
		\includegraphics[width=\textwidth]{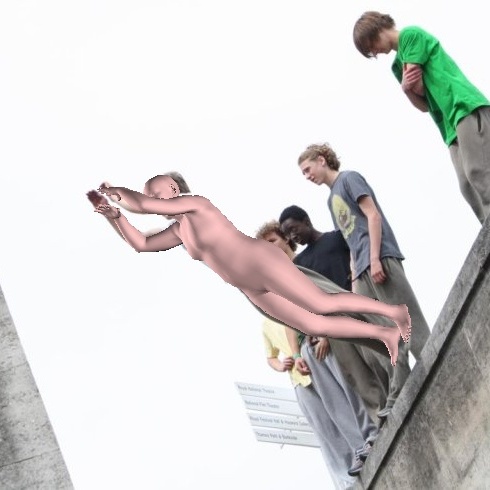}
	\end{subfigure}\\
	\begin{subfigure}[]{.15\textwidth}
		\includegraphics[width=\textwidth]{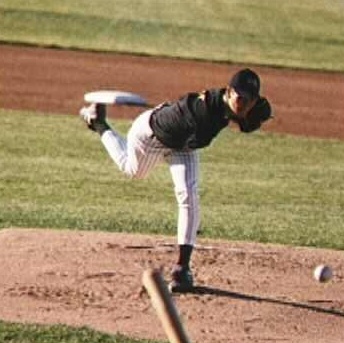}
	\end{subfigure}~
	\begin{subfigure}[]{.15\textwidth}
		\includegraphics[width=\textwidth]{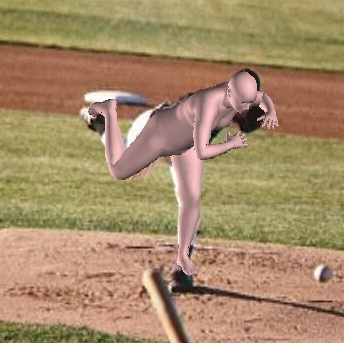}
	\end{subfigure}~
	\begin{subfigure}[]{.15\textwidth}
		\includegraphics[width=\textwidth]{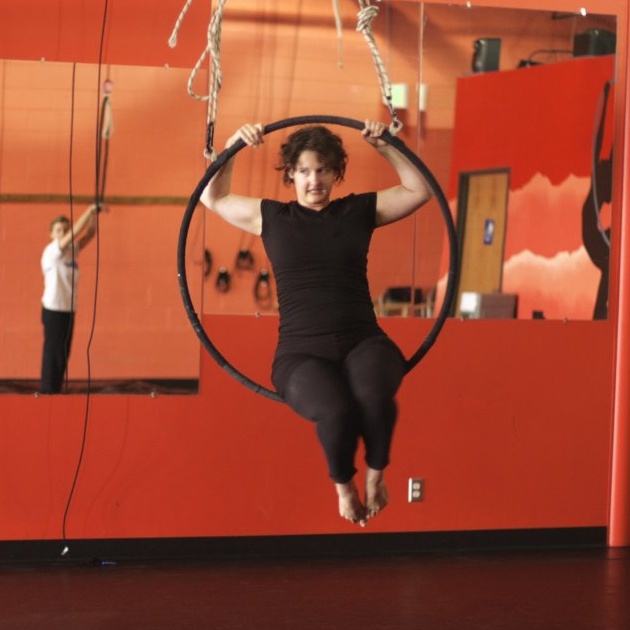}
	\end{subfigure}~
	\begin{subfigure}[]{.15\textwidth}
		\includegraphics[width=\textwidth]{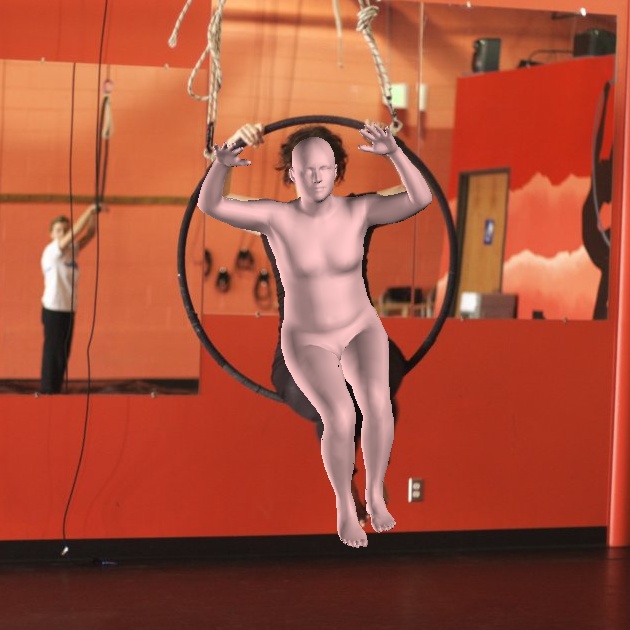}
	\end{subfigure}~
	\begin{subfigure}[]{.15\textwidth}
		\includegraphics[width=\textwidth]{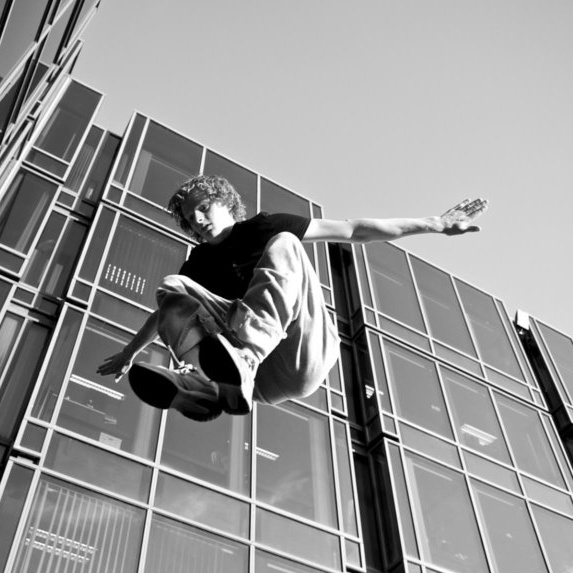}
	\end{subfigure}~
	\begin{subfigure}[]{.15\textwidth}
		\includegraphics[width=\textwidth]{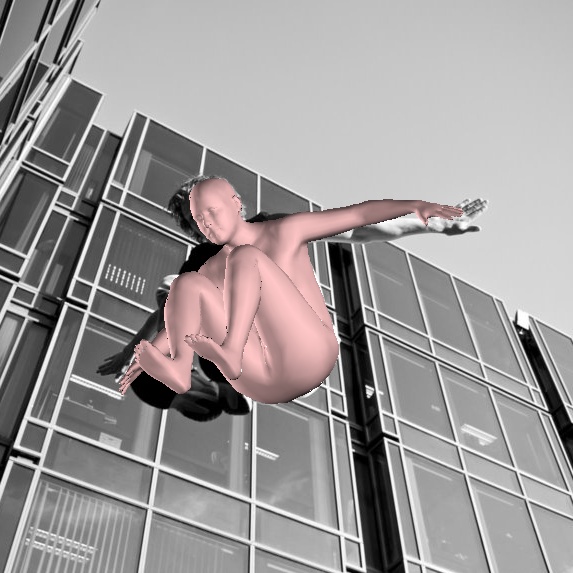}
	\end{subfigure}\\
		\begin{subfigure}[]{.15\textwidth}
		\includegraphics[width=\textwidth]{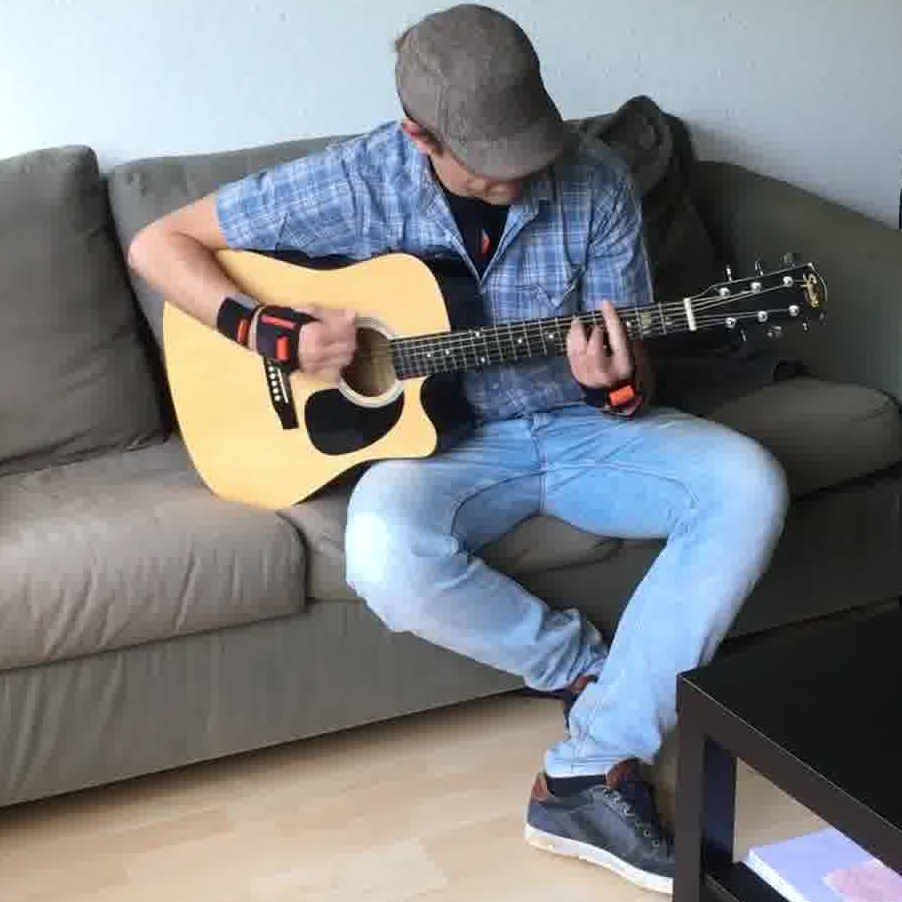}
	\end{subfigure}~
	\begin{subfigure}[]{.15\textwidth}
		\includegraphics[width=\textwidth]{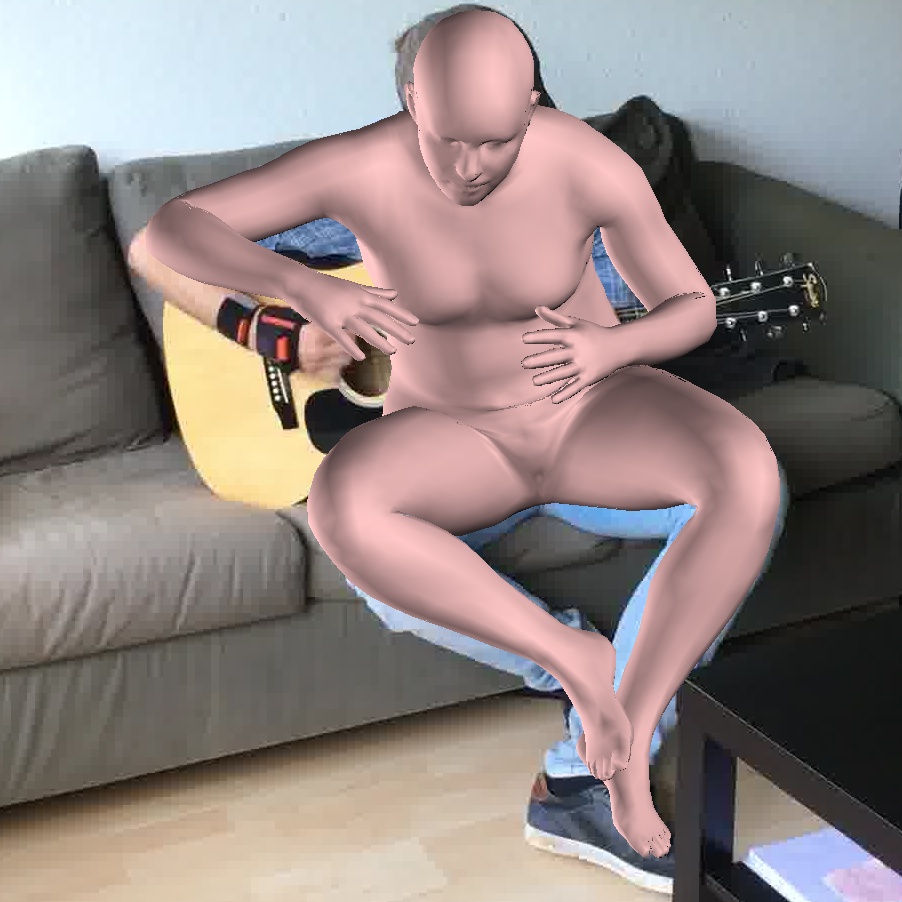}
	\end{subfigure}~
	\begin{subfigure}[]{.15\textwidth}
		\includegraphics[width=\textwidth]{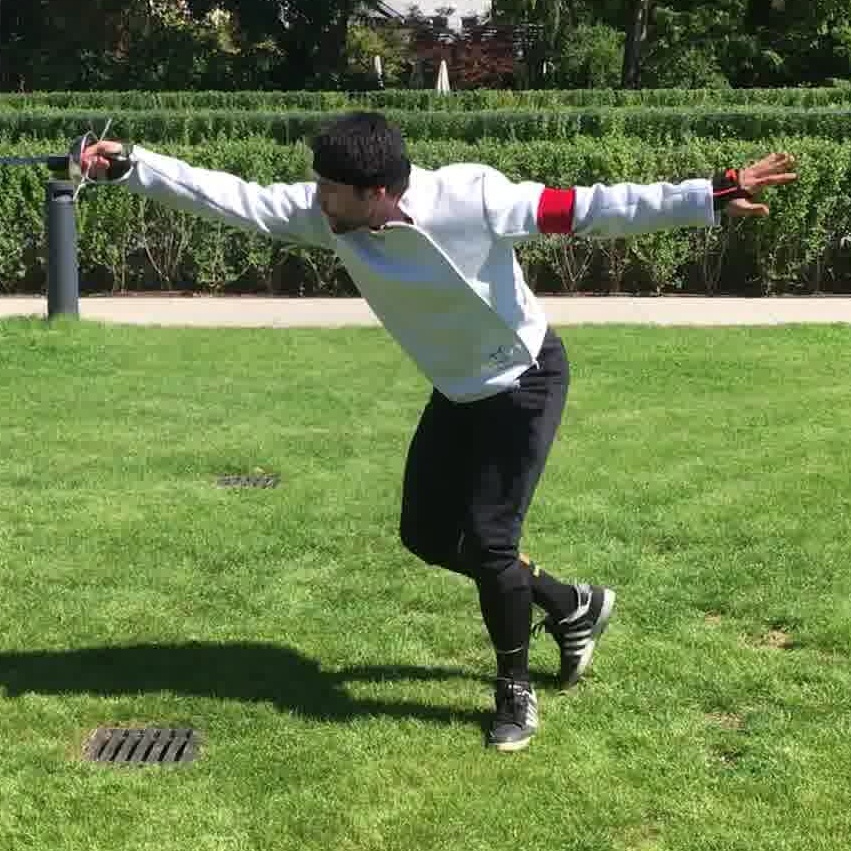}
	\end{subfigure}~
	\begin{subfigure}[]{.15\textwidth}
		\includegraphics[width=\textwidth]{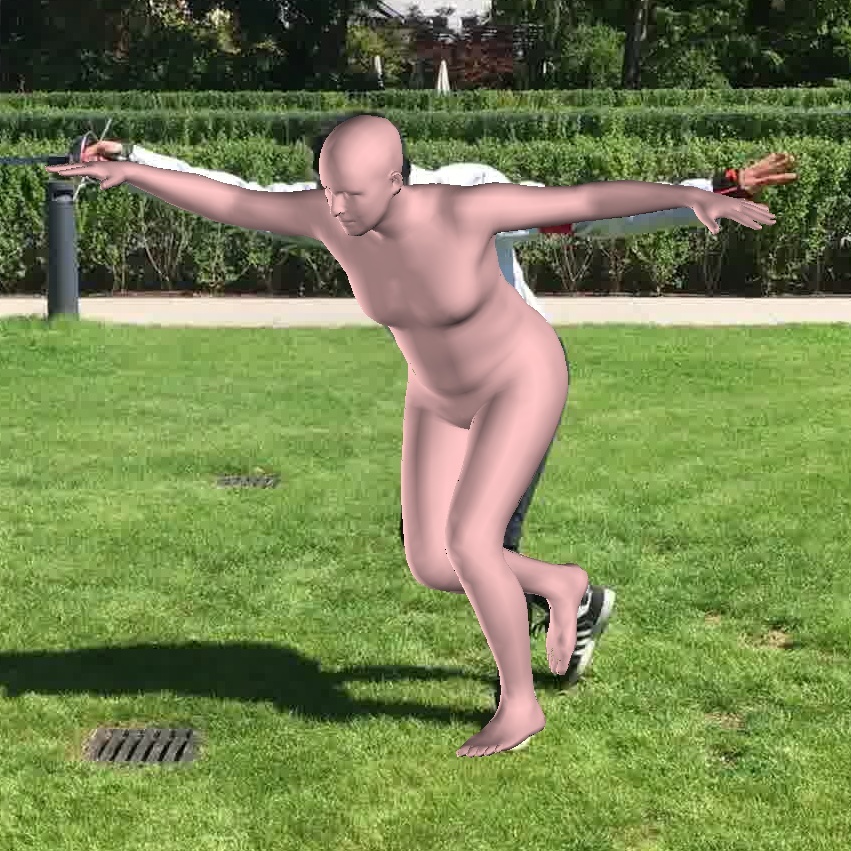}
	\end{subfigure}~
	\begin{subfigure}[]{.15\textwidth}
		\includegraphics[width=\textwidth]{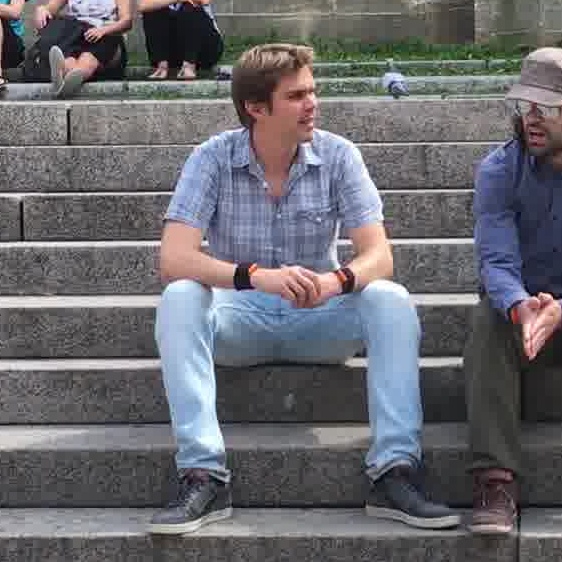}
	\end{subfigure}~
	\begin{subfigure}[]{.15\textwidth}
		\includegraphics[width=\textwidth]{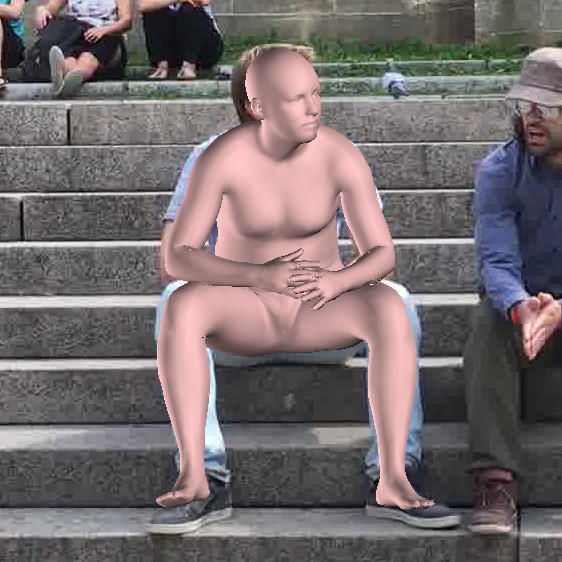}
	\end{subfigure}\\
	\begin{subfigure}[]{.15\textwidth}
		\includegraphics[width=\textwidth]{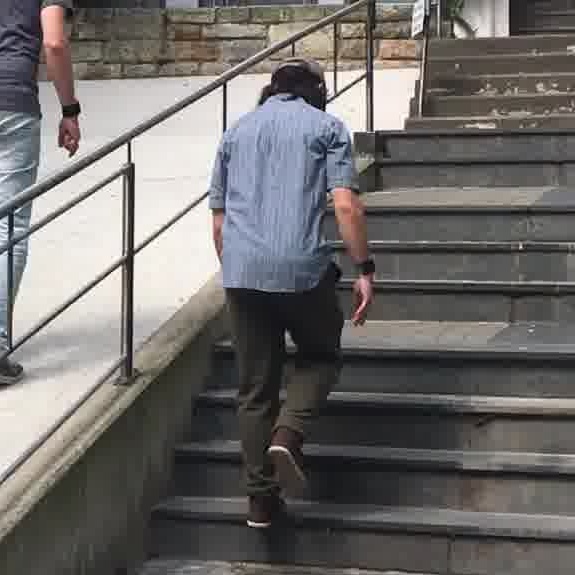}
	\end{subfigure}~
	\begin{subfigure}[]{.15\textwidth}
		\includegraphics[width=\textwidth]{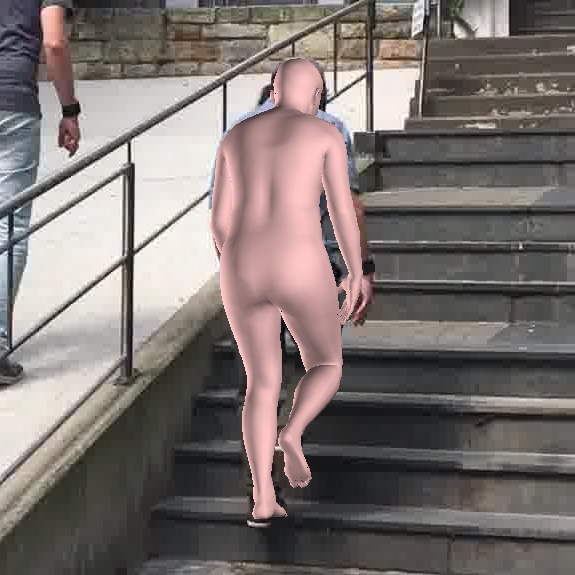}
	\end{subfigure}~
	\begin{subfigure}[]{.15\textwidth}
		\includegraphics[width=\textwidth]{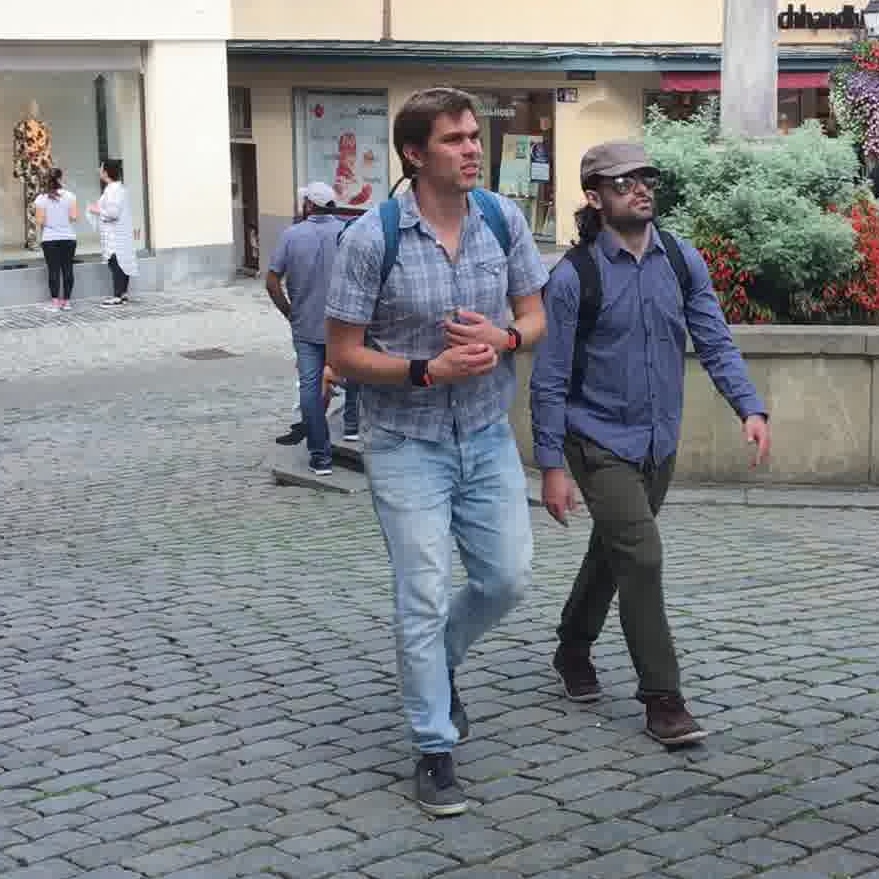}
	\end{subfigure}~
	\begin{subfigure}[]{.15\textwidth}
		\includegraphics[width=\textwidth]{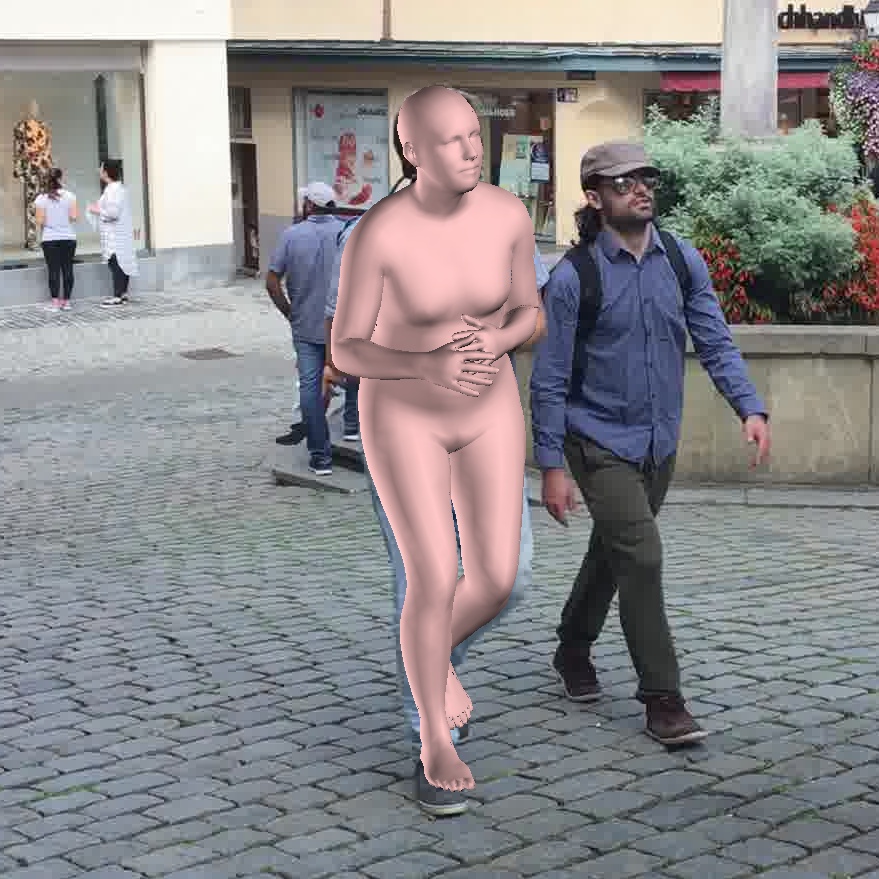}
	\end{subfigure}~
	\begin{subfigure}[]{.15\textwidth}
		\includegraphics[width=\textwidth]{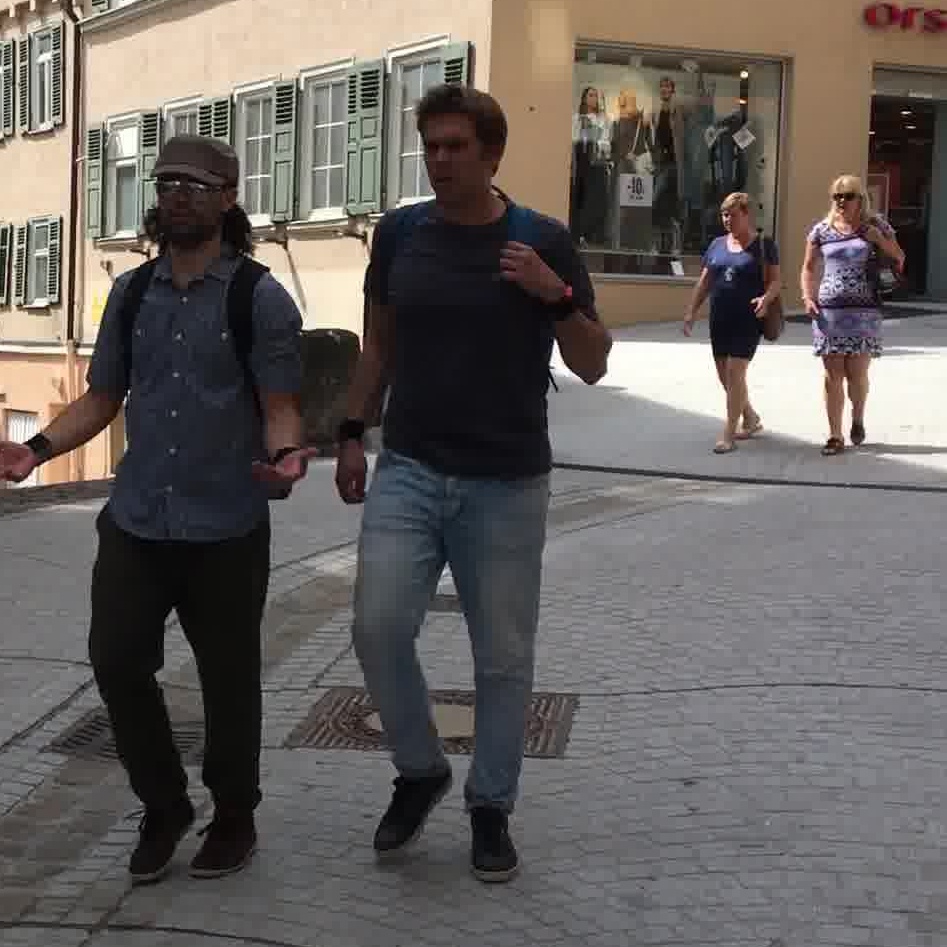}
	\end{subfigure}~
	\begin{subfigure}[]{.15\textwidth}
		\includegraphics[width=\textwidth]{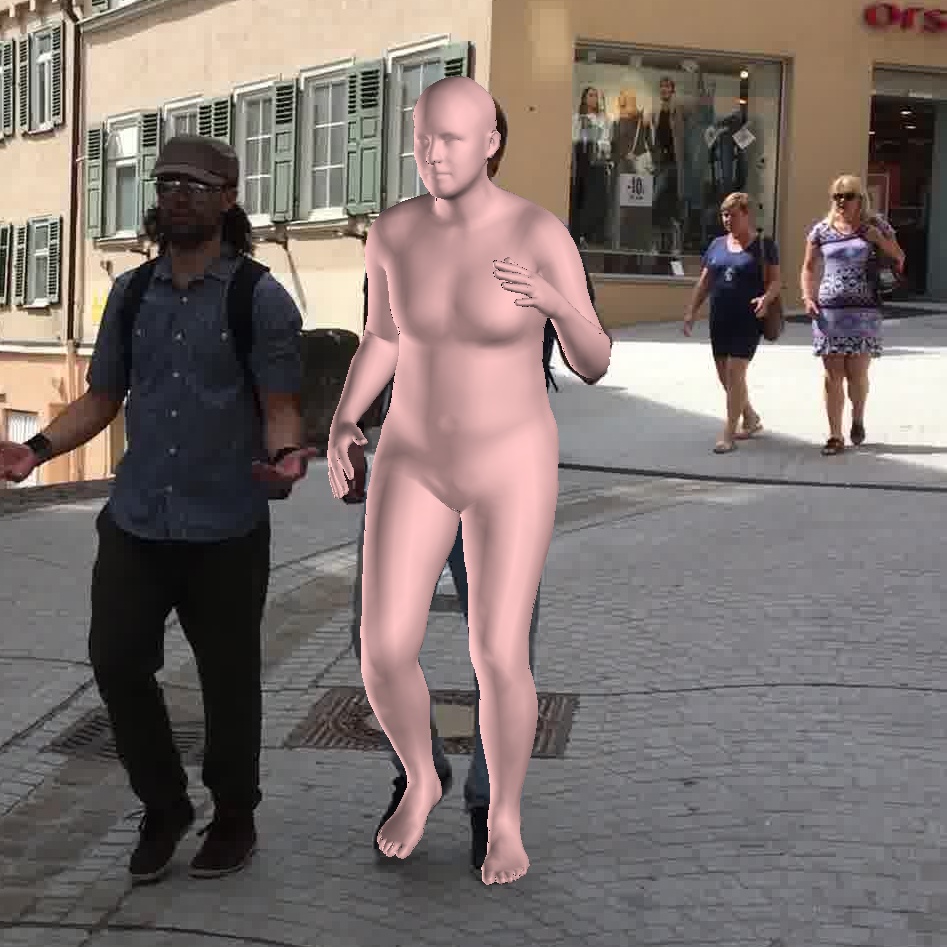}
	\end{subfigure}\\
	\begin{subfigure}[]{.15\textwidth}
		\includegraphics[width=\textwidth]{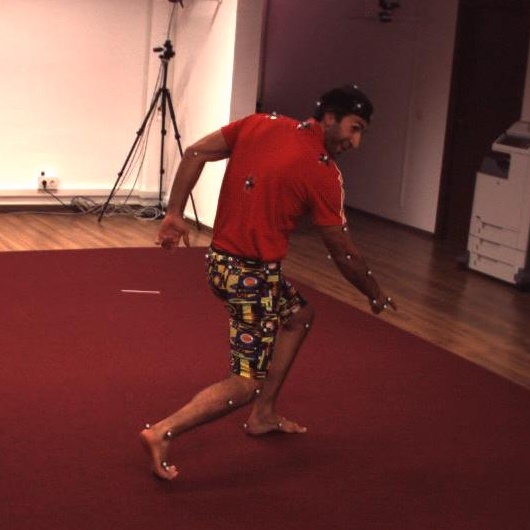}
	\end{subfigure}~
	\begin{subfigure}[]{.15\textwidth}
		\includegraphics[width=\textwidth]{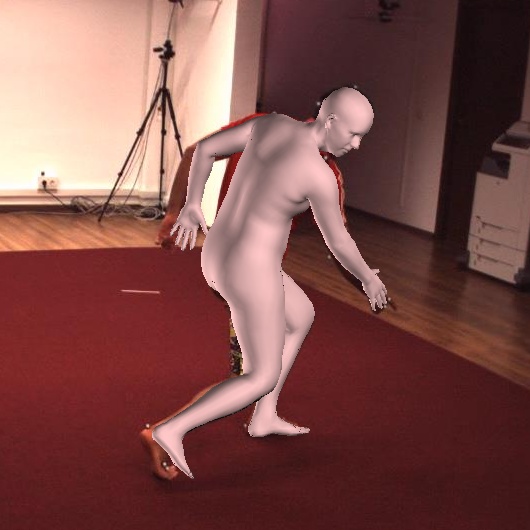}
	\end{subfigure}~
	\begin{subfigure}[]{.15\textwidth}
		\includegraphics[width=\textwidth]{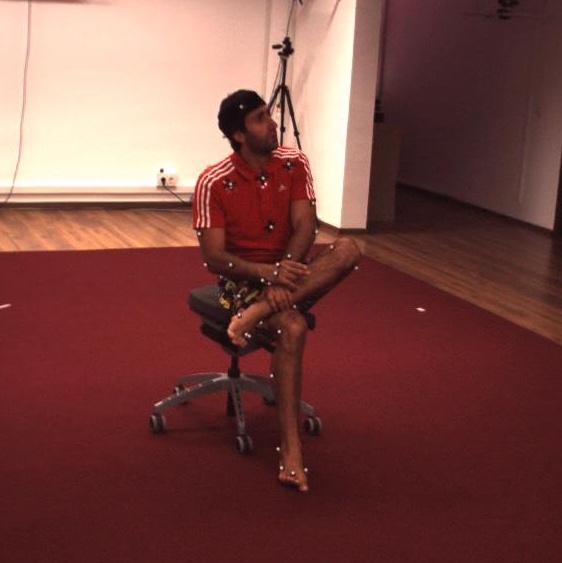}
	\end{subfigure}~
	\begin{subfigure}[]{.15\textwidth}
		\includegraphics[width=\textwidth]{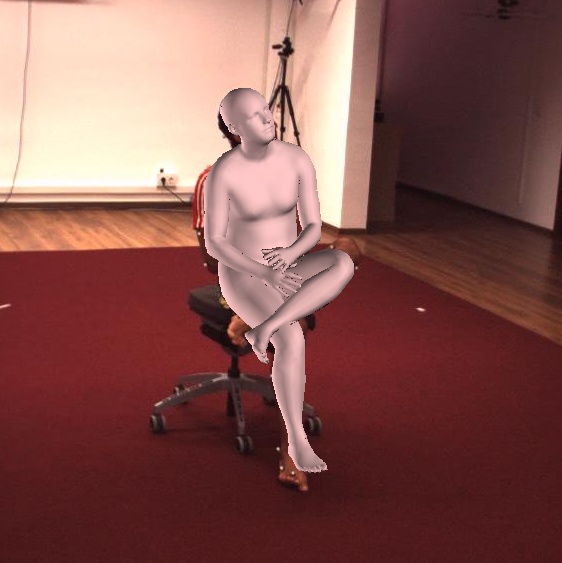}
	\end{subfigure}~
	\begin{subfigure}[]{.15\textwidth}
		\includegraphics[width=\textwidth]{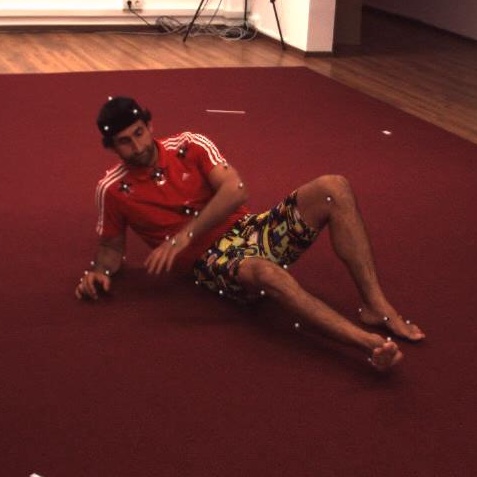}
	\end{subfigure}~
	\begin{subfigure}[]{.15\textwidth}
		\includegraphics[width=\textwidth]{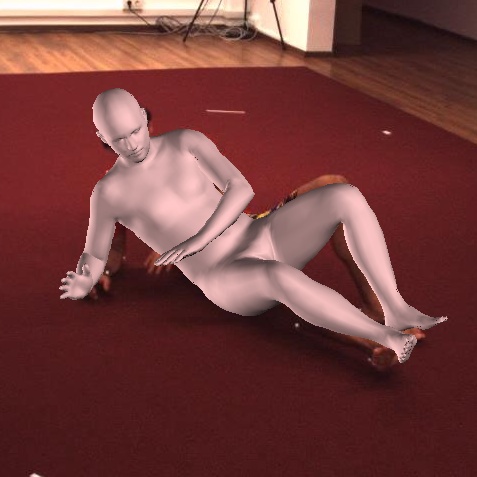}
	\end{subfigure}\\
	\begin{subfigure}[]{.15\textwidth}
		\includegraphics[width=\textwidth]{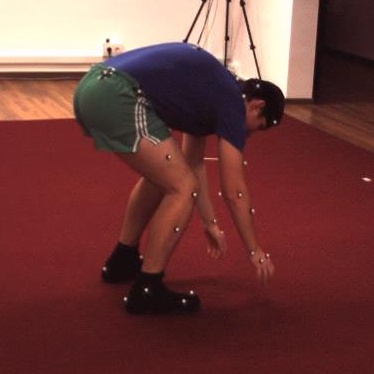}
	\end{subfigure}~
	\begin{subfigure}[]{.15\textwidth}
		\includegraphics[width=\textwidth]{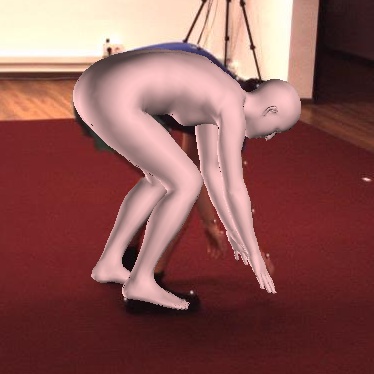}
	\end{subfigure}~
	\begin{subfigure}[]{.15\textwidth}
		\includegraphics[width=\textwidth]{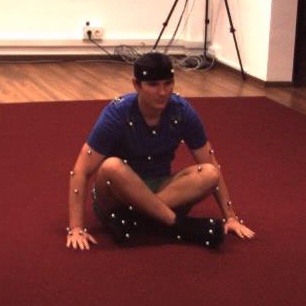}
	\end{subfigure}~
	\begin{subfigure}[]{.15\textwidth}
		\includegraphics[width=\textwidth]{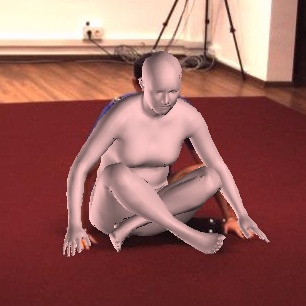}
	\end{subfigure}~
	\begin{subfigure}[]{.15\textwidth}
		\includegraphics[width=\textwidth]{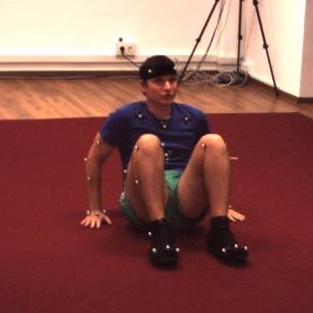}
	\end{subfigure}~
	\begin{subfigure}[]{.15\textwidth}
		\includegraphics[width=\textwidth]{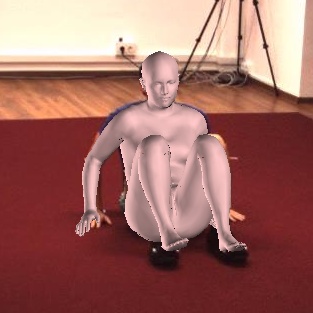}
	\end{subfigure}\\
	\begin{subfigure}[]{.15\textwidth}
		\includegraphics[width=\textwidth]{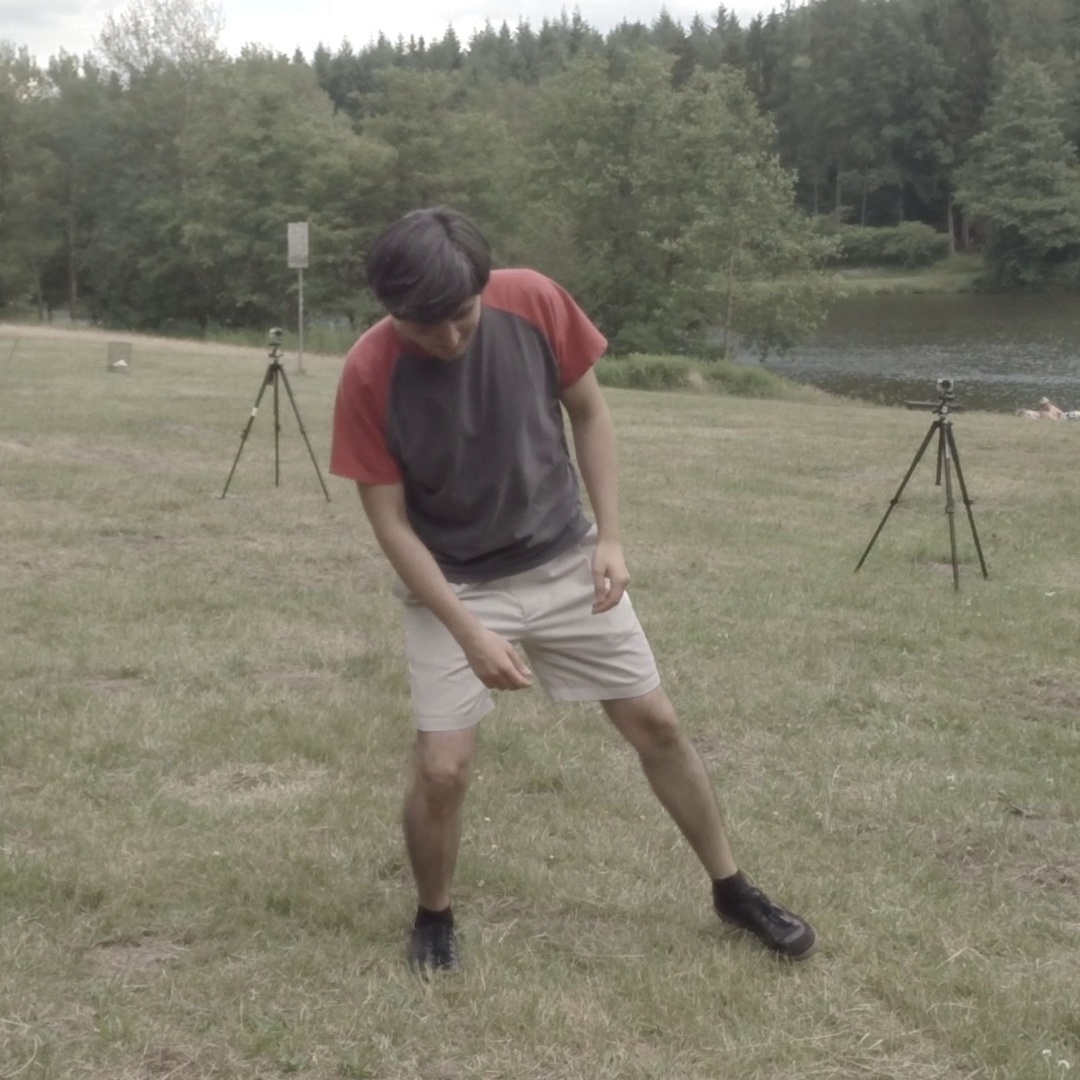}
	\end{subfigure}~
	\begin{subfigure}[]{.15\textwidth}
		\includegraphics[width=\textwidth]{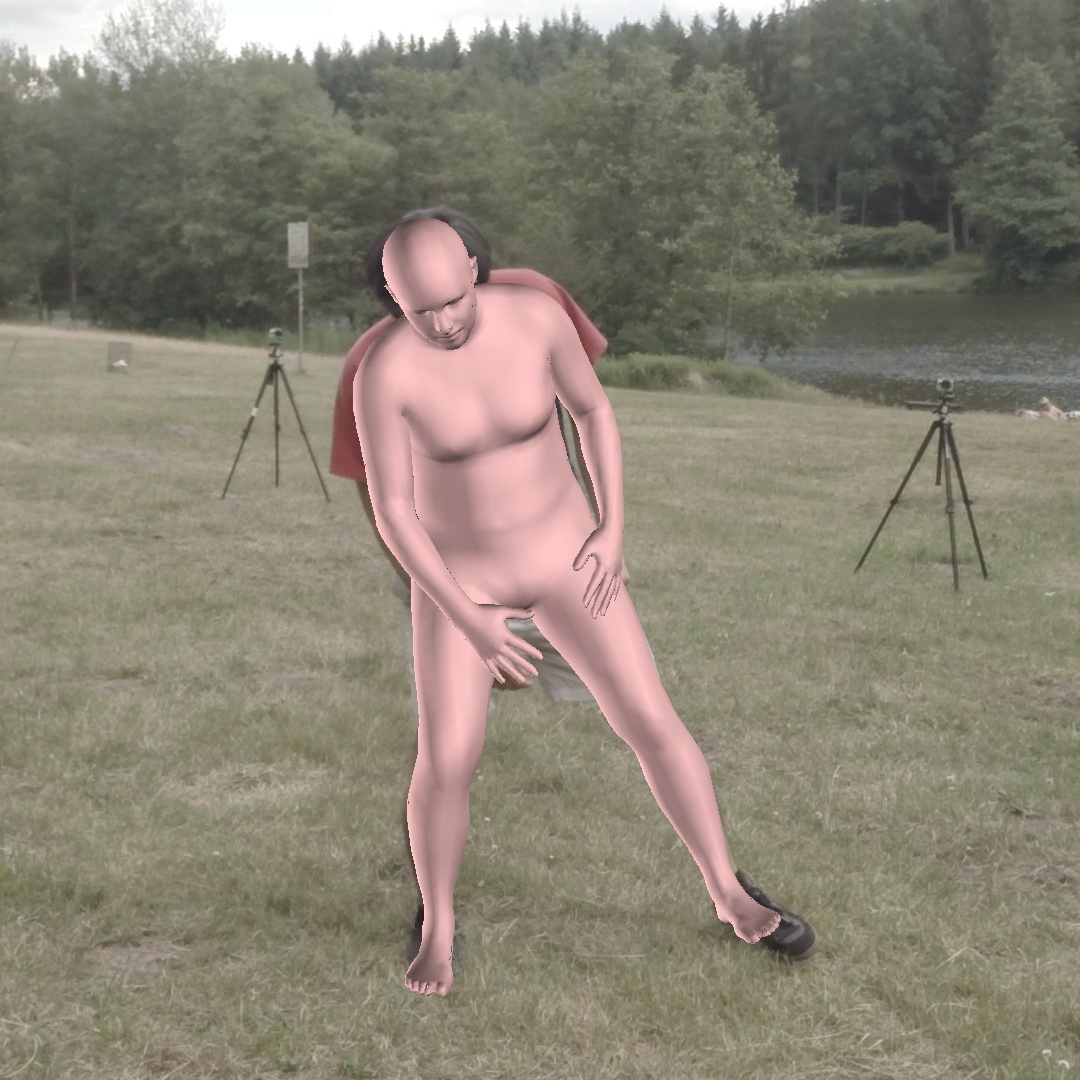}
	\end{subfigure}~
	\begin{subfigure}[]{.15\textwidth}
		\includegraphics[width=\textwidth]{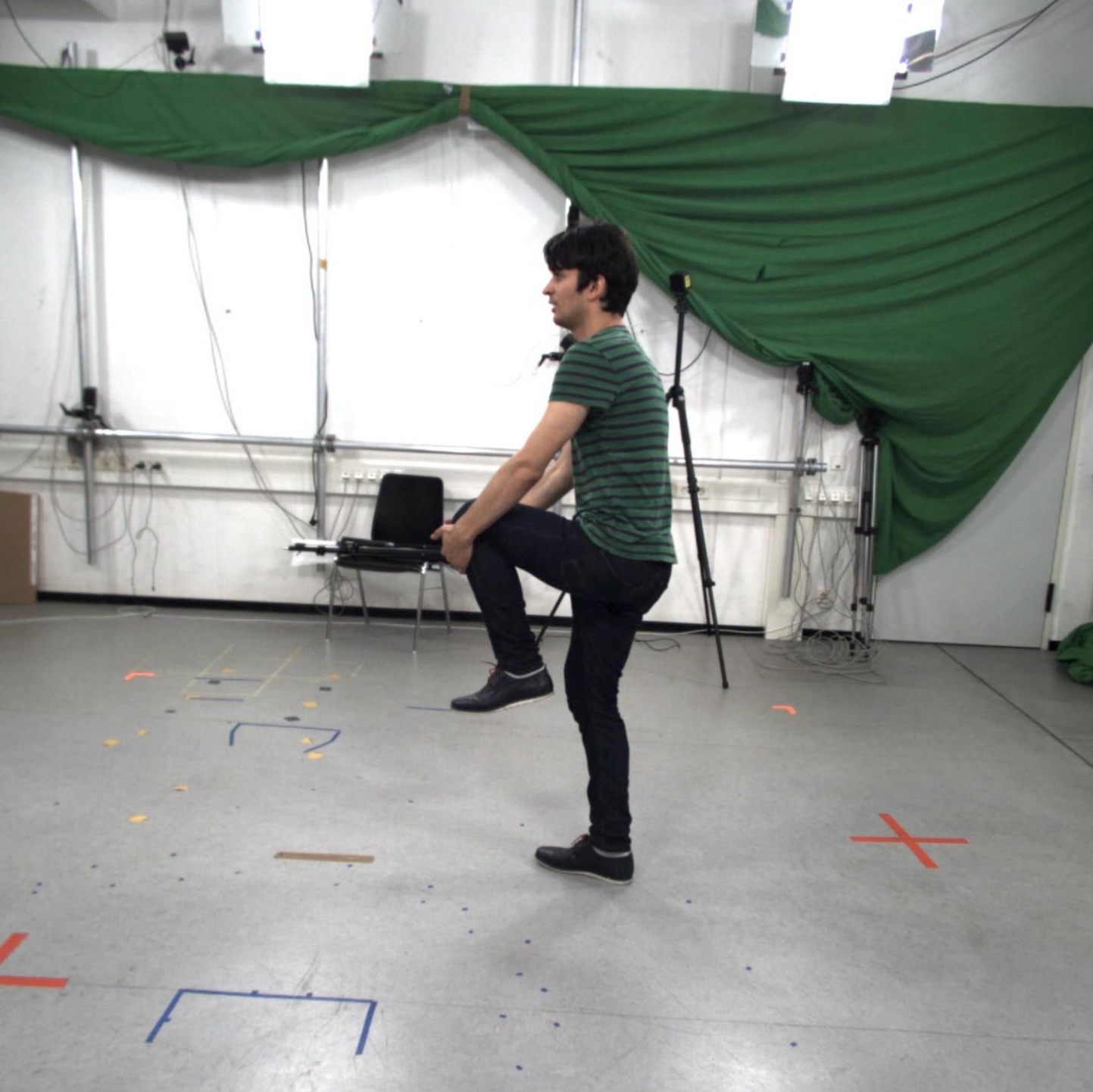}
	\end{subfigure}~
	\begin{subfigure}[]{.15\textwidth}
		\includegraphics[width=\textwidth]{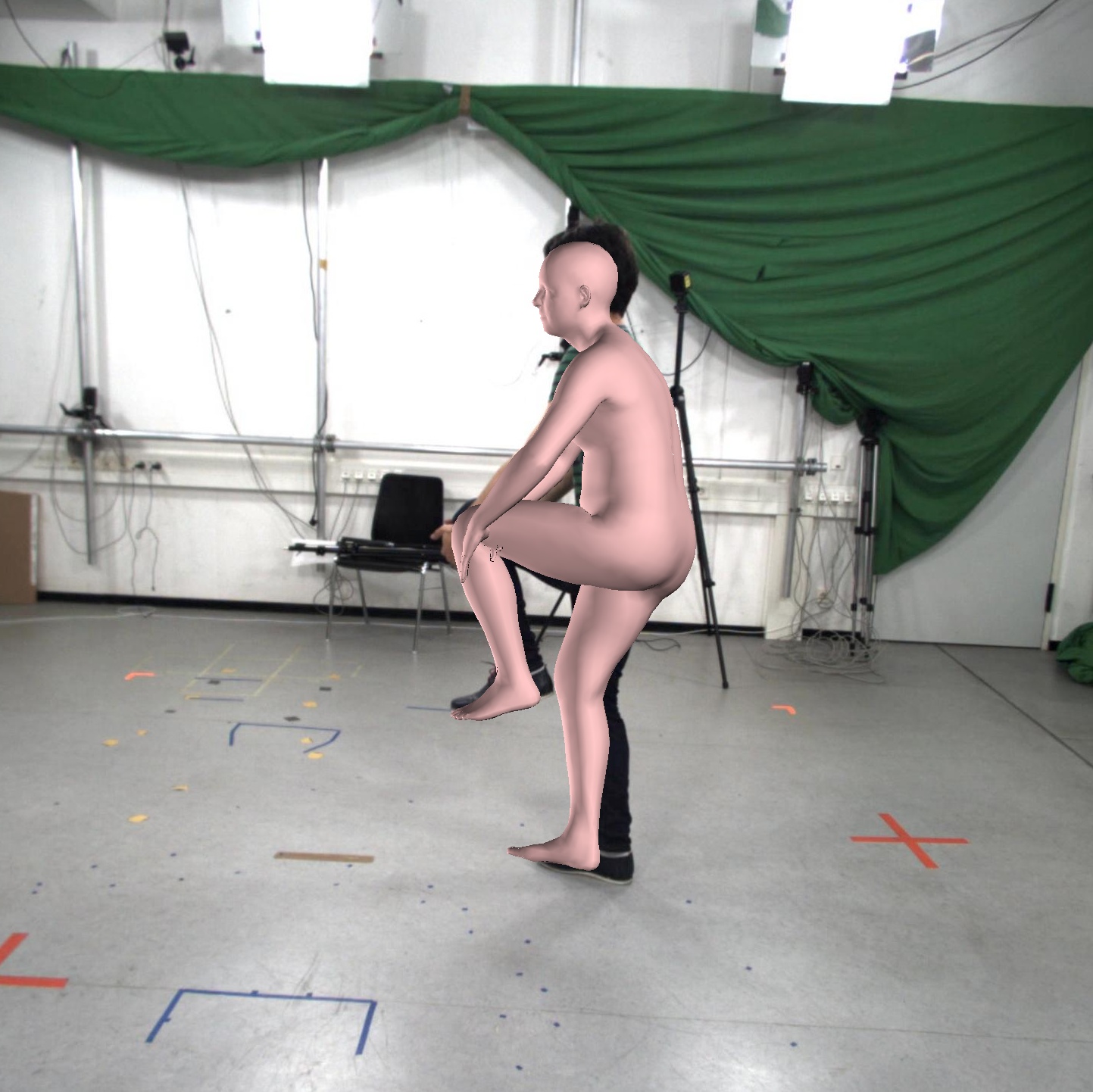}
	\end{subfigure}~
	\begin{subfigure}[]{.15\textwidth}
		\includegraphics[width=\textwidth]{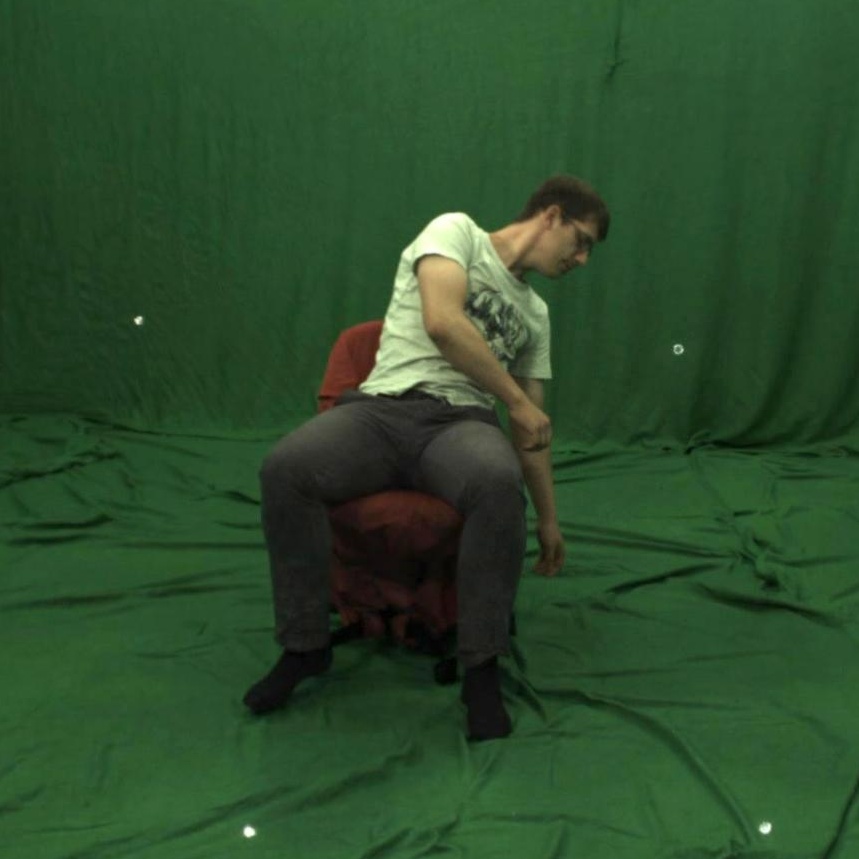}
	\end{subfigure}~
	\begin{subfigure}[]{.15\textwidth}
		\includegraphics[width=\textwidth]{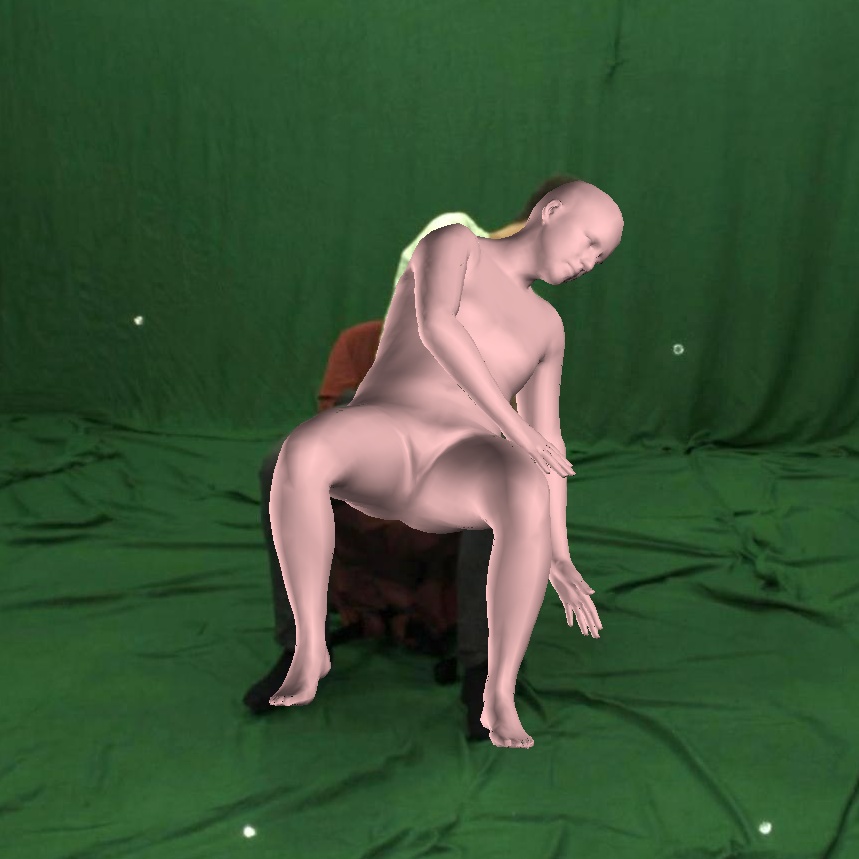}
	\end{subfigure}\\
\vspace{-2mm}
	\caption{Qualitative results from various datasets, LSP (rows 1-3), 3DPW (rows 4-5), H36M (rows 6-7) and MPI-INF-3DHP (row 8).}
\label{fig:results}
\vspace{-2mm}
\end{figure*}

The same comparison is performed for the LSP dataset. In this case, we evaluate 3D shape implicitly through mesh reprojection and evaluation of silhouette and part segmentation accuracy. The full results for this setting are presented in Table~\ref{tab:lsp}. The trend here is similar to the 3DPW results. Using a static set of fits and providing model-based supervision achieves very compelling results. However, it is the incorporation of the optimization in the loop that propels our approach beyond the state-of-the-art.

To better illustrate the degree of improvement for fits in our dictionary, we provide some typical examples in Figure~\ref{fig:comparison}. As the training progresses, the fits improve significantly, giving to the network access to better supervision.

\textbf{Comparison with the state-of-the-art}:
For further comparison with the state-of-the-art, we report results in additional datasets for 3D human pose estimation. Based on the different settings, proposed in the literature, we report results both when we use 3D ground truth whenever it is available (e.g.,~Human3.6M), and also when no image with 3D ground truth is available for training. Similarly to~\cite{kanazawa2018end}, we call this setting ``unpaired'', since images and 3D ground truth do not come in pairs for training.

\begin{table}
\centering
\small
\hspace{-3mm}
\tabcolsep=0.85mm
\begin{tabular}{@{}lc@{}}
\toprule
& Rec. Error \\
\midrule
Lassner~\etal~\cite{lassner2017unite} & 93.9 \\
SMPLify~\cite{bogo2016keep} & 82.3 \\
Pavlakos~\etal~\cite{pavlakos2018learning} & 75.9 \\
HMR (unpaired)~\cite{kanazawa2018end} & 66.5 \\
Ours (unpaired) & \bf{62.0} \\
\midrule
NBF~\cite{omran2018neural} & 59.9 \\
HMR~\cite{kanazawa2018end} & 56.8 \\
Ours & \bf{41.1} \\
\bottomrule
\end{tabular}
\vspace{-2mm}
\caption{Evaluation on the Human3.6M dataset. The numbers are mean reconstruction errors in mm. We compare with approaches that output a mesh of the human body. Approaches on the top part require no image with 3D ground truth, while approaches on the bottom part make use of 3D ground truth too. In both settings, our approach outperforms the state-of-the-art by significant margins.}
\label{tab:h36m}
\vspace{-2mm}
\end{table}

In Table~\ref{tab:h36m}, we present the results of our approach on Human3.6M against other approaches that also output a full mesh of the human body (SMPL, in particular). Our approach outperforms the previous baselines when 3D ground truth is not available for training (top of the table) and when it is (bottom). We highlight that for the case that no 3D ground truth is available (e.g., unpaired setting), our network does not have access to poses from Human3.6M as Kanazawa~\etal~\cite{kanazawa2018end}, since our pose prior is trained only on CMU data. Despite that, we still outperform~\cite{kanazawa2018end}.

Similarly, we also report results on the MPI-INF-3DHP dataset, for the two settings (paired/unpaired supervision). Again, we outperform~\cite{kanazawa2018end}, while being very competitive against two approaches that do not use a parametric model of the human body~\cite{mehta2017monocular,mehta2017vnect}.

\begin{table}
\centering
\small
\hspace{-3mm}
\tabcolsep=0.85mm
\begin{tabular}{@{}lcccccc@{}}
\toprule
& \multicolumn{3}{c}{Absolute} & \multicolumn{3}{c}{Rigid Alignment} \\
\cmidrule{2-7}
& PCK & AUC & MPJPE & PCK & AUC & MPJPE \\
\midrule
HMR (unpaired)~\cite{kanazawa2018end} & 59.6 & 27.9  & 169.5 & 77.1 & 40.7 & 113.2 \\
Ours (unpaired) & \textbf{66.8} & \textbf{30.2} & \textbf{124.8} & \textbf{87.0} & \textbf{48.5} & \textbf{80.4} \\
\midrule
Mehta~\etal~\cite{mehta2017monocular} & 75.7 & 39.3 & 117.6 & - & - & - \\
VNect~\cite{mehta2017vnect} & \textbf{76.6} & \textbf{40.4} & 124.7 & 83.9 & 47.3 & 98.0 \\
HMR~\cite{kanazawa2018end} & 72.9 & 36.5 & 124.2 & 86.3 & 47.8 & 89.8 \\
Ours & 76.4 & 37.1 & \textbf{105.2} & \textbf{92.5} & \textbf{55.6} & \textbf{67.5} \\
\bottomrule
\end{tabular}
\vspace{-2mm}
\caption{Evaluation on the MPI-INF-3DHP dataset. The comparison is under different metrics before (left) and after (right) rigid alignment. Our approach outperforms the previous baselines. (For PCK and AUC, higher is better, while for MPJPE, lower is better).}
\label{tab:3dhp}
\vspace{-2mm}
\end{table}

\begin{figure}[!htb]
	\centering
	\begin{subfigure}[]{.2\columnwidth}
		\includegraphics[width=\textwidth]{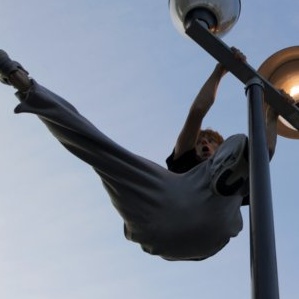}
	\end{subfigure}~
	\begin{subfigure}[]{.2\columnwidth}
		\includegraphics[width=\textwidth]{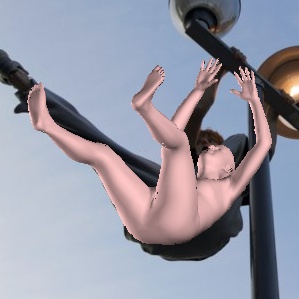}
	\end{subfigure}~
	\begin{subfigure}[]{.2\columnwidth}
		\includegraphics[width=\textwidth]{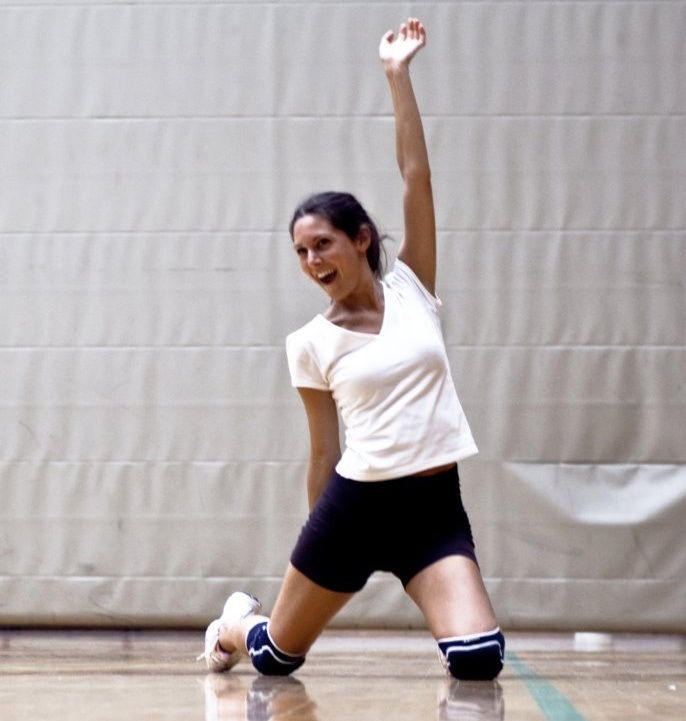}
	\end{subfigure}~
	\begin{subfigure}[]{.2\columnwidth}
		\includegraphics[width=\textwidth]{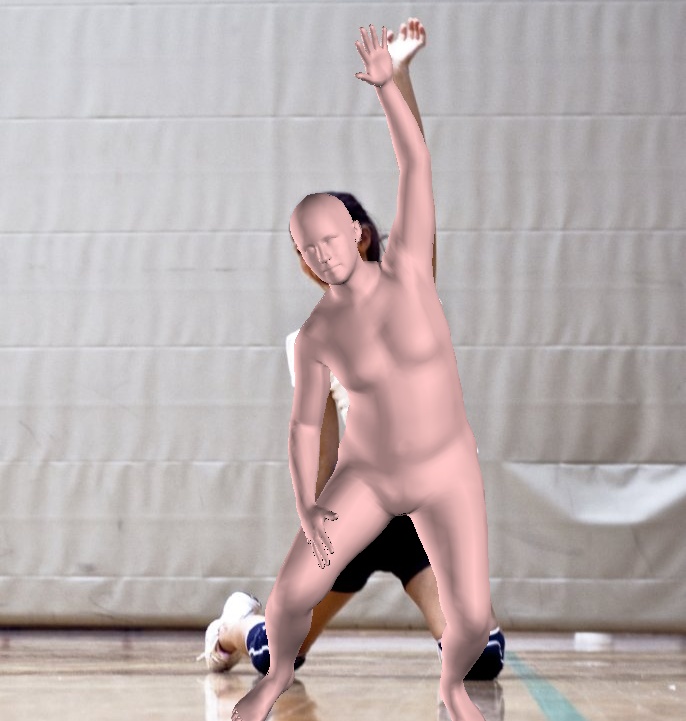}
	\end{subfigure}\\
	\begin{subfigure}[]{.2\columnwidth}
		\includegraphics[width=\textwidth]{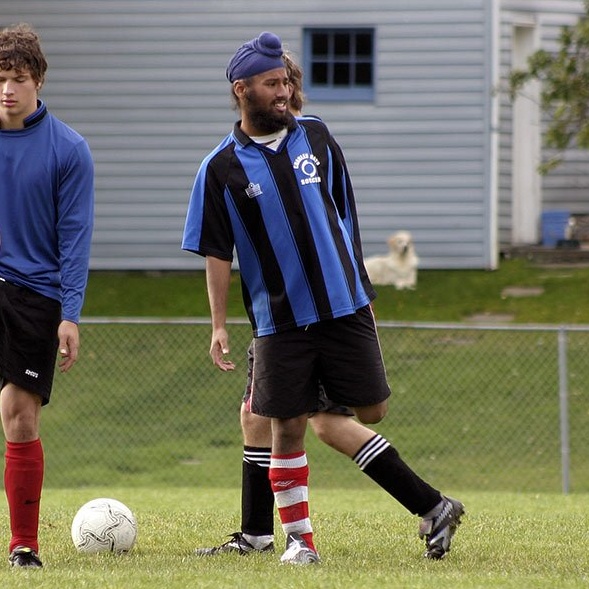}
			\caption*{Image}	
	\end{subfigure}~
	\begin{subfigure}[]{.2\columnwidth}
		\includegraphics[width=\textwidth]{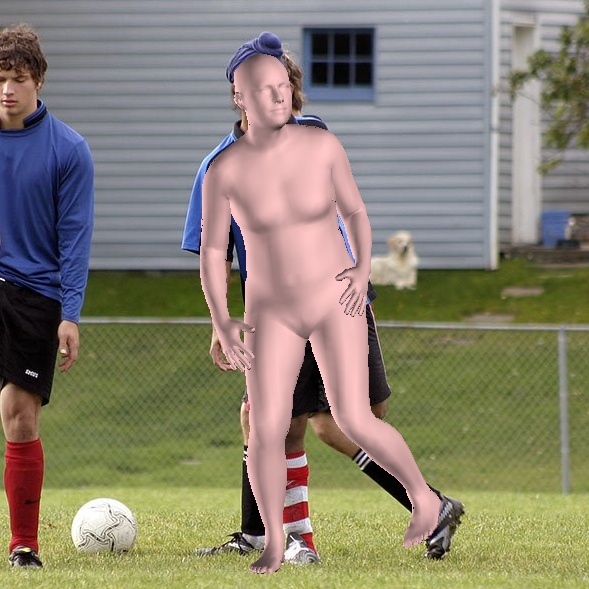}
			\caption*{Result}			
	\end{subfigure}~
	\begin{subfigure}[]{.2\columnwidth}
		\includegraphics[width=\textwidth]{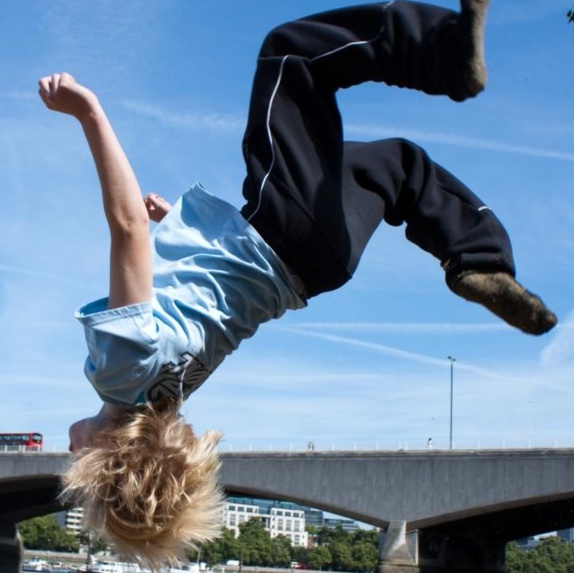}
			\caption*{Image}			
	\end{subfigure}~
	\begin{subfigure}[]{.2\columnwidth}
		\includegraphics[width=\textwidth]{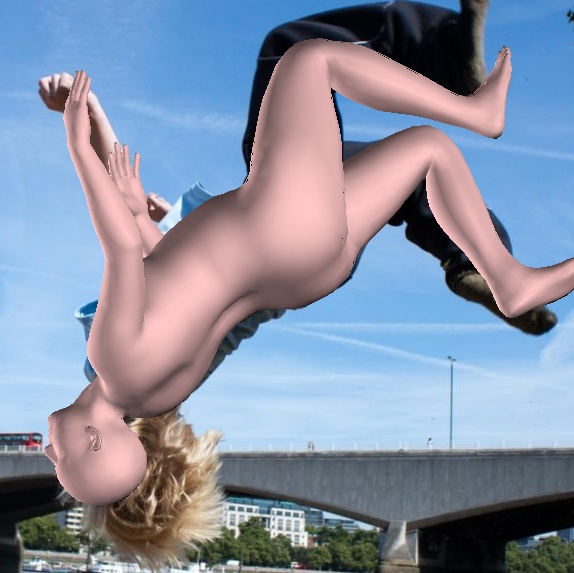}
			\caption*{Result}			
	\end{subfigure}
	\caption{Erroneous reconstructions of our network. Typical failure cases can be attributed to challenging poses, ordinal depth ambiguities, viewpoints which are rare in the training set, as well as confusion due to the existence of multiple people in the scene.}
\label{fig:failures}	
\end{figure}

Finally, Figure~\ref{fig:results} includes qualitative results of our approach from the different datasets involved in our evaluation, while Figure~\ref{fig:failures} includes some failure cases. A larger variety of results can also be found in the Sup.Mat.

\section{Summary}
This work describes \spin, an approach that proposes a close collaboration between a regression method and an optimization-based method to train a deep network for 3D human pose and shape estimation. Our approach uses the network to provide an initial estimate to the optimization routine, which then fits the model in the loop and provides model-based supervision for the training of the network. Thus, the optimization-module and regression-module form a self-improving cycle since they can both benefit through their tight collaboration. Moreover, the privileged model-based supervision is valuable to improve the training of our network, which is also demonstrated by the empirical results, where our approach outperforms previous approaches by large margins. Simultaneously, since the fitting routine requires only 2D keypoints to fit the model, we can train our deep network even in the absence of 3D annotations. Future work could consider extending this approach to capture multiple people~\cite{zanfir2018monocular,zanfir2018deep}, or incorporate more expressive models of the human body~\cite{joo2018total,pavlakos2019expressive}.

\footnotesize
\noindent
{\bf Acknowledgements:} NK, GP and KD gratefully appreciate support through the following grants: 
NSF-IIP-1439681 (I/UCRC), NSF-IIS-1703319, NSF MRI 1626008, ARL RCTA W911NF-10-2-0016, ONR N00014-17-1-2093, ARL DCIST CRA W911NF-17-2-0181, the DARPA-SRC C-BRIC, by Honda Research Institute and a Google Daydream Research Award.

\vspace{+01.00mm}
\footnotesize
\noindent
{\bf Disclosure:} 
MJB has received research gift funds from Intel, Nvidia, Adobe, Facebook, and Amazon. 
While MJB is a part-time employee of Amazon, his research was performed solely at, and funded solely by, MPI. 
MJB has financial interests in Amazon and Meshcapade GmbH.

{\small
\balance
\bibliographystyle{ieee_fullname}
\bibliography{egbib}
}

\end{document}